\PassOptionsToPackage{hyphens}{url}
\documentclass[9pt]{article}

\usepackage{textcomp}
\usepackage[margin=1in]{geometry}
\usepackage[numbers,sort&compress]{natbib}
\setcitestyle{square}
\usepackage[english]{babel}
\usepackage{gensymb}
\usepackage{svg}
\usepackage{graphicx}
\DeclareGraphicsExtensions{.png,.pdf}
\usepackage{float}
\usepackage{subcaption}
\usepackage{amsmath}
\usepackage{array}
\usepackage{booktabs}
\usepackage{adjustbox}
\usepackage{algorithm}
\usepackage{algpseudocode}
\usepackage{xcolor}
\usepackage{ragged2e}
\usepackage[outdir=./]{epstopdf}
\usepackage{longtable}
\usepackage{xurl}
\usepackage[hidelinks,breaklinks]{hyperref}
\usepackage{xltabular}
\newcolumntype{Y}{>{\raggedright\arraybackslash}X}
\usepackage{orcidlink}
\usepackage{tikz}
\usetikzlibrary{arrows.meta,positioning}

\title{CROWD: A global dataset of continuous urban dashcam driving}

\author{%
\begin{tabular}{c}
\textbf{Md Shadab Alam}$^{1,*}$\,\orcidlink{0000-0001-9184-9963},\;
\textbf{Olena Bazilinska}$^{2}$\,\orcidlink{0000-0001-6239-6766},\;
\textbf{Pavlo Bazilinskyy}$^{1}$\,\orcidlink{0000-0001-9565-8240}
\\[1.2ex]
$^{1}$Eindhoven University of Technology, Eindhoven, The Netherlands\\
$^{2}$National University of Kyiv-Mohyla Academy, Kyiv, Ukraine
\\[1.2ex]
$^{*}$Corresponding author: Md Shadab Alam (\nolinkurl{m.s.alam@tue.nl})
\end{tabular}
}

\date{}
\begin{document}
\maketitle

\begin{abstract}
We introduce CROWD (City Road Observations With Dashcams), a manually curated dataset of ordinary, variable duration, temporally contiguous, unedited, front facing urban dashcam segments screened and segmented from publicly available YouTube videos. CROWD is designed to support cross-domain robustness and interaction analysis by prioritising routine driving and explicitly excluding crashes, crash aftermath, and other edited or incident-focused content. The release contains 107,407 segment records spanning 36,204.26 hours (82,484 unique uploads), covering 10,538 named inhabited places in 238 countries and territories across all six inhabited continents (Africa, Asia, Europe, North America, South America and Oceania), with segment level manual labels for time of day (day or night) and vehicle type. To lower the barrier for reproducible baseline analysis, we provide per-segment CSV files of automatic detections for all 80 MS-COCO classes produced with YOLOv11x, together with segment-local multi-object tracks generated with BoT-SORT; e.g. person, bicycle, motorcycle, car, bus, truck, traffic light, stop sign, etc. CROWD is distributed as video identifiers with segment boundaries and derived detection and tracking pseudo-labels, enabling reproducible research without redistributing the underlying videos. To support leakage-free evaluation, we also provide canonical splits grouped by source channel or by locality, and weights that balance the contribution of localities and countries.

\end{abstract}

\section{Background \& Summary}
\label{sec:background}

Urban traffic is a visually and behaviourally complex environment in which vulnerable road users (VRUs), particularly pedestrians and cyclists, interact with motor vehicles under substantial variation in infrastructure, lighting, road geometry, and culturally specific driving norms. The World Health Organisation estimates 1.19 million deaths from road traffic per year, with pedestrians accounting for 23\% and cyclists for 6\% of fatalities worldwide \cite{WHO2023RoadSafety}. Automated transport research therefore benefits from datasets that capture VRU appearance and motion across diverse urban settings, rather than within a narrow set of locations or carefully curated scenarios.

Police reported collision statistics, such as the UK's STATS19 system~\cite{DfTStats19}, and national crash databases, such as the US Fatality Analysis Reporting System (FARS)~\cite{NHTSAFARS}, provide essential and standardised measures of crash outcomes and circumstances. STATS19 captures around 50 coded data elements per injury collision (e.g.\ time and location, vehicle types and manoeuvres, and basic driver and casualty attributes), while FARS codes more than 170 data elements per fatal crash spanning crash level, vehicle and driver level, and person level variables (e.g.\ injury severity, restraint use, and alcohol involvement). However, these resources are typically released as \emph{tabular} records, meaning that each crash (and its associated vehicles and people) is represented as rows of coded variables rather than as continuous, time aligned video or other sensor streams. They are also centred on the crash event and its immediate circumstances. Consequently, they rarely preserve the continuous visual context needed to study how interactions unfold over time, how occlusion and scene clutter affect detection, or how routine exposure relates to risk. Naturalistic driving studies address part of this limitation by instrumenting volunteer vehicles to record extended video from the driver's viewpoint together with vehicle state over long periods, but they are expensive to conduct and often constrained by privacy and access restrictions, which limits reuse and benchmarking at scale~\cite{Dingus2006_100Car,antin2019second, eenink2014udrive,TRB_SHRP2_SafetyDataAccess}.

Curated benchmarks for driving perception, spanning general street scene driving datasets such as KITTI~\cite{geiger2012we}, Cityscapes~\cite{cordts2016cityscapes}, ApolloScape~\cite{huang2018apolloscape}, BDD100K~\cite{yu2020bdd100k}, Mapillary Vistas~\cite{neuhold2017mapillary}, KITTI 360~\cite{liao2022kitti}, and A2D2~\cite{geyer2020a2d2}, as well as autonomous vehicle focussed multi sensor datasets such as Argoverse~\cite{chang2019argoverse}, Argoverse 2~\cite{wilson2023argoverse}, nuScenes~\cite{caesar2020nuscenes}, the Waymo Open Dataset~\cite{sun2020scalability}, and PandaSet~\cite{xiao2021pandaset}, have accelerated methodological advances by providing high quality sensor data and annotations. These benchmarks are fundamental for detection, segmentation, tracking, and forecasting, but they also reflect practical constraints of collection. Geographic coverage is often limited to a small number of cities or regions, many releases prioritise short selected scenarios rather than long uninterrupted driving, and fleets instrumented with specialised sensor suites do not necessarily reflect the viewpoint, mounting geometry, or image characteristics of widely deployed camera based driver assistance systems. VRU-focused datasets further support pedestrian and cyclist modelling, but are also geographically bounded. EuroCity Persons is explicitly European, recorded in 31 cities across 12 European countries~\cite{braun2018eurocity}. PIE was recorded in the centre of Toronto, ON, Canada~\cite{rasouli2019pie}. JAAD was recorded in a limited set of locations in North America and Europe~\cite{kotseruba2016joint}. Models trained on one dataset can generalise poorly to new data distributions, reinforcing the need for diverse and ecologically valid data to reduce dataset bias and domain shift~\cite{torralba2011unbiased}.

Publicly available web video offers an alternative route to scale and diversity, especially via dashcam recordings. However, web shared dashcam content is frequently shaped by selection effects. Clips that are posted and widely circulated disproportionately feature unusual or high conflict events, including collisions, near misses, and confrontations, which skews the observed distribution away from routine driving. This bias is visible in dashcam benchmarks that focus on accidents and critical events~\cite{fang2019dada}. For research questions that require representative samples of everyday urban driving, including typical VRU exposure, everyday interactions, and background traffic flow, there remains a gap for geographically diverse data deliberately filtered towards ordinary continuous driving in urban settings.

A dataset captured from a forward facing dashcam provides advantages for both method development and behavioural analysis (dashcam cameras may be mounted inside the vehicle, for example behind the windscreen, or externally, for example on the roof or bonnet. Some externally mounted setups can produce footage that looks very similar to an in vehicle dashcam because of the camera angle and settings). It matches the egocentric viewpoint used by many deployed camera based pipelines and it foregrounds perceptual challenges that dominate real deployments, including partial occlusion by parked vehicles and street furniture, dense roadside clutter, frequent scale changes as VRUs approach the ego vehicle, and complex intersection geometries that produce ambiguous motion cues. Dashcams are widespread in many regions and are typically mounted in broadly comparable windscreen positions, enabling scalable collection across regions while maintaining a coherent observational geometry for cross locality and cross country comparisons. Throughout this paper, we use the terms \emph{country} and countries to refer to countries and territories as defined by ISO 3166, including dependent territories, represented using ISO 3166 1 alpha 3 codes (field \texttt{iso3}) \cite{iso3166}. However, prevalence and motivations are country specific, with dashcams used mainly for safety and evidential purposes in some contexts and more often for leisure recording and online sharing in others.

The CROWD (City Road Observations With Dashcams) dataset addresses these needs by curating extended duration, front facing urban dashcam video segments sourced from publicly available online recordings, with an emphasis on ordinary continuous driving. Videos containing crashes, crash aftermath, or other incident-focused content are excluded by design. The dataset prioritises routine urban conditions across multiple countries and localities and, to our knowledge, provides the largest geographic coverage among publicly available curated datasets of ordinary, temporally contiguous urban dashcam driving. It also offers a large total retained duration, particularly within this broad geographic scope. CROWD supports studies of robustness and domain generalisation, as well as analyses that depend on temporal continuity within retained compliant intervals, such as multi object tracking, exposure estimation, and interaction characterisation. Alongside video identifiers, segment definitions, and dataset metadata, CROWD provides machine generated object bounding boxes and segment-local tracks aligned to video frames as derived baseline outputs. These outputs are intended as automatic pseudo-labels that can support reproducible baselines and follow on experiments in detection, tracking, and interaction analysis without requiring users to rerun the full detection pipeline; they should not be interpreted as manually validated ground truth.

\begin{figure}[H]
  \includegraphics[width=\textwidth]{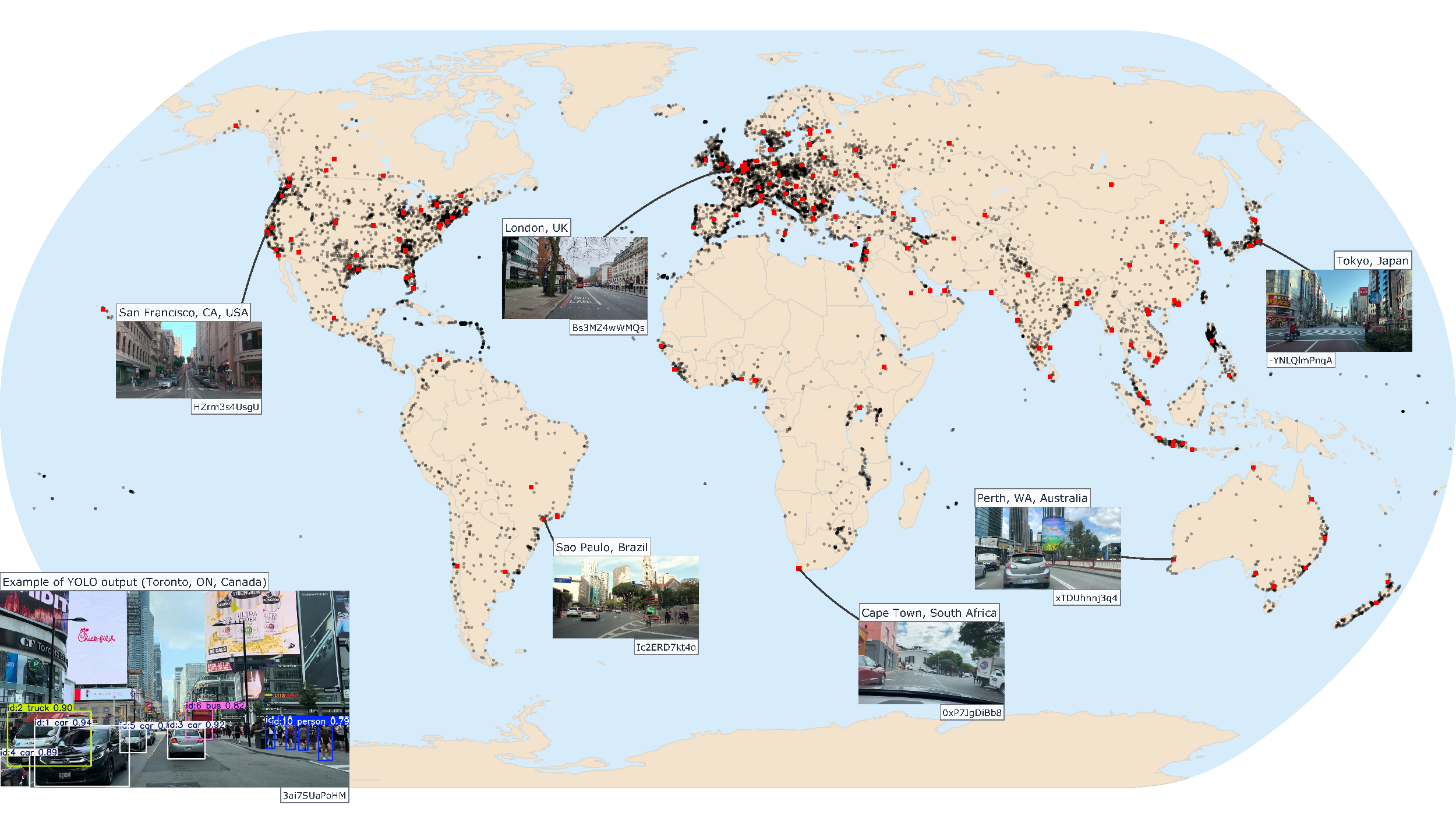}
  \caption{Geographic coverage of CROWD across 10,538 localities worldwide. Each marker denotes one locality record with at least one retained dashcam segment. Marker positions are approximate locality-level coordinates, obtained from the named locality, optional state or region, and country, with manual adjustment where needed during curation. They do not represent route coordinates, GPS traces, exact recording locations, segment start points, or frame-level positions. Red markers indicate localities with more than 100,000 s of retained footage (n = 158); all other localities are shown with the base marker style. Insets show selected example localities linked to their approximate locality positions on the map. Geolocation uncertainty is therefore at the locality scale, and the coordinates should not be interpreted as street-level ground truth.}
  \label{fig:cities_map}
\end{figure}


\subsection{Related datasets}
\label{sec:dataset}

A substantial body of prior work has released driving video datasets to support perception, tracking, and scene understanding. Although early driving video resources date back to CamVid~\cite{brostow2009semantic} and the Caltech Pedestrian Dataset~\cite{dollar2009pedestrian}, large scale dataset releases have accelerated over the last decade as in vehicle cameras have become inexpensive, storage has become cheaper, and online video platforms have made recording and sharing at scale more accessible. This shift has been reinforced by the widespread use of dashcams for evidential purposes in collisions and disputes, including insurance related claims and fraud attempts, and by sharing clips to seek help from others or to support reporting via trusted authorities such as the police, although concerns about privacy, retaliation, and criminal misuse can shape what is uploaded and retained~\cite{park2016motives, kim2020dashcam}. These datasets vary in objective, acquisition pipeline, sensor suite, geographic coverage, and the extent to which they capture urban routine driving as continuous video sequences rather than isolated frames. \autoref{tab:dashcam_datasets} summarises representative resources that provide a video stream mounted on the vehicle in the forward direction, offering a viewpoint that is broadly comparable to consumer dashcam footage even when the sensor package or mounting differs.

Several of the resources summarised in \autoref{tab:dashcam_datasets} are distributed as clips or log segments on the order of tens of seconds, for example 20\,s scenes in nuScenes~\cite{caesar2020nuscenes} and 15 to 30\,s sequences in Argoverse~\cite{chang2019argoverse}. Such durations are well suited to benchmarking per frame detection, segmentation, and related tasks, but offer limited continuity for analyses that require longer continuous observation, such as interaction and yielding behaviour over extended encounters, exposure estimation expressed per hour of driving, and cross national behavioural measures derived from event timing in naturalistic footage~\cite{alam2025pedestrian}. BDD100K, for instance, provides 100,000 front facing 40\,s clips with diversity in weather and time of day, yet the fixed clip length constrains longer horizon interaction analysis and exposure oriented measures~\cite{yu2020bdd100k}. D$^{2}$-City similarly provides urban dashcam clips with substantial scenario diversity, but the data released are segmented into short excerpts \cite{che2019d}. Comparable design choices appear in short log releases from multisensor platforms, where the forward camera stream is available but the temporal window per scene is brief, as in nuScenes, Waymo Open Dataset, Argoverse 2 Sensor, and PandaSet~\cite{caesar2020nuscenes,sun2020scalability,wilson2023argoverse,xiao2021pandaset}. These datasets are invaluable for model development and evaluation, while their temporal granularity can limit studies that require uninterrupted driving context.

\begingroup
\setlength{\tabcolsep}{5pt} 
\renewcommand{\arraystretch}{1.15}
\footnotesize
\emergencystretch=3em 

\begin{xltabular}{\textwidth}{@{}p{0.10\textwidth} p{0.22\textwidth} p{0.10\textwidth} p{0.12\textwidth} X@{}}
\caption{Representative driving datasets reported in the literature that provide a forward facing camera stream. For multi sensor datasets, the comparison refers to the forward camera stream or streams only. The cited papers are publicly accessible. Dataset access may be subject to registration, approval, or a licence agreement, and availability can change over time. Footage duration is reported in hours when explicitly stated by the source. Otherwise, the closest published proxy, such as distance or frame counts, is listed and indicated in the notes below the table.}\label{tab:dashcam_datasets}\\
\toprule
\textbf{Dataset} & \textbf{Geographic coverage (countries/localities)} & \textbf{Footage (h)} & \textbf{Day/Night} & \textbf{Notes (dashcam video only)} \\
\midrule
\endfirsthead

\multicolumn{5}{@{}l}{\footnotesize \textbf{Table \thetable\ (continued).}}\\
\toprule
\textbf{Dataset} & \textbf{Geographic coverage (countries/localities)} & \textbf{Footage (h)} & \textbf{Day/Night} & \textbf{Notes (dashcam video only)} \\
\midrule
\endhead

\midrule
\multicolumn{5}{r@{}}{\footnotesize Continued on next page.}\\
\endfoot

\bottomrule
\multicolumn{5}{@{}p{\textwidth}@{}}{\vspace{3pt}\footnotesize Access status checked on 4 March 2026. We could verify a public access route for 23 of the 26 datasets listed at that date, via open download, registration, or acceptance of a licence agreement. IDD-CRS is currently listed as private on India Data, HDD requires a university affiliated download request, and ZOD requires requesting access by email. The access status of the four datasets added in October 2026 (Argoverse 1, Cityscapes, Mapillary Vistas and UDRIVE) was not checked in the same way.  $^{a}$Hours of driving data collected reported as 47,382.65 h in the Phase II report.   $^{b}$4Seasons is reported primarily by distance and sequences rather than total hours.   $^{c}$A2D2 is reported via sequential frame counts across cities.   $^{d}$1000$\times$(15 to 20) s = 4.17 to 5.56 h.   $^{e}$100,000$\times$40 s = 1,111.11 h.   $^{f}$Boreas is reported primarily by distance (350+ km) rather than total hours.   $^{g}$CADC is reported primarily by frame counts (for example 7000 frames annotated) rather than total hours.   $^{h}$74$\times$5 min = 6.17 h.   $^{i}$KITTI 360 is published primarily via distance and frame counts, rather than total hours (for example 73.7 km).   $^{j}$nuPlan reports 1,200 h of human driving data.   $^{k}$1000$\times$20 s = 5.56 h.   $^{l}$Oxford RobotCar is reported as distance and images rather than total hours.   $^{m}$103$\times$8 s = 0.23 h.   $^{n}$2030$\times$20 s = 11.28 h.   $^{o}$1473$\times$20 s = 8.18 h (sequences only).   $^{p}$113$\times$(15 to 30) s = 0.47 to 0.94 h; the motion forecasting data of Argoverse 1 (from 1,006 driving hours) has no camera images.   $^{q}$Cityscapes is reported as frame counts rather than total hours.   $^{r}$Mapillary Vistas consists of still images, so no footage duration applies.   $^{s}$More than 88,000 h of vehicle data are reported by the project (final report); the hours of camera footage are not itemised.   $^{t}$2,019 segments of 1 min = 33.65 h.}\\
\endlastfoot

100 Car Naturalistic Driving Study (Phase II)~\cite{Dingus2006_100Car} & USA (Northern Virginia, Metro Washington, DC) & 47,382.65$^{a}$ & Both & Instrumented vehicles with five camera views including a forward view; infrared cabin lighting supports night driving capture; event database includes crashes, near crashes, and incidents; a public release is reported for an event subset (time series and annotations). \\

4Seasons~\cite{wenzel20204seasons} & Germany (locations not itemised) & N/A$^{b}$ & Both & Vehicle mounted forward view available; reported primarily by distance rather than total hours. \\

A2D2~\cite{geyer2020a2d2} & Germany (Gaimersheim, Ingolstadt, Munich) & N/A$^{c}$ & Not specified & Multi sensor dataset; sequential camera frames are available (reported as 392,556 frames across three cities). \\

A3CarScene\linebreak \cite{cantarini2023a3carscene} & Italy (Marche region) & 31 & Not specified & Two dashcams mounted on front and rear windows; recorded on public roads (audio also provided). \\

Argoverse 1 (3D Tracking)~\cite{chang2019argoverse} & USA (Pittsburgh, PA; Miami, FL) & 0.47 to 0.94$^{p}$ & Not specified & 113 log segments of 15 to 30 s with 7 ring cameras and a forward facing stereo pair; multi sensor logs. \\

Argoverse 2 (Sensor)~\cite{wilson2023argoverse} & USA (6 cities: Austin, TX; Detroit, MI; Miami, FL; Palo Alto, CA; Pittsburgh, PA; Washington, DC) & 4.17 to 5.56$^{d}$ & Both & 1,000 short logs (15 to 20 s) with surround cameras; forward facing stereo pair available; multi sensor logs. \\

BDD100K~\cite{yu2020bdd100k} & USA (New York, NY; San Francisco Bay Area, CA; other regions not specified) & 1,111.11$^{e}$ & Both & 100,000 front view videos; each is $\sim$40 s; includes diverse weather and times of day (daytime and nighttime). \\

Boreas~\cite{burnett2023boreas} & Canada (Toronto, ON) & N/A$^{f}$ & Both & Multi sensor dataset; forward camera stream available; reported primarily by distance ($>$350 km). \\

CADC~\cite{pitropov2021canadian} & Canada (Region of Waterloo, ON) & N/A$^{g}$ & Not specified & Adverse winter driving dataset; forward camera streams available (8 cameras); reported primarily by frame counts. \\

Caltech Pedestrians~\cite{dollar2009pedestrian} & USA (Los Angeles, CA) & 10 & Not specified & Forward facing video recorded from a moving vehicle in urban traffic; annotated pedestrian bounding boxes with occlusion labels; available via CaltechDATA. \\

Cityscapes~\cite{cordts2016cityscapes} & Germany and neighbouring countries (50 cities) & N/A$^{q}$ & Day & Stereo video recorded in spring, summer and fall, without adverse weather; 5,000 finely and 20,000 coarsely annotated frames, each with a 30-frame snippet. \\

comma2k19\linebreak \cite{schafer2018commute} & USA (CA\textendash 280 between San Jose and San Francisco, CA) & 33.65$^{t}$ & Not specified & Road facing camera; 2,019 segments of 1 minute each. \\

D$^{2}$\textendash City~\cite{che2019d} & China (6 cities, not itemised in the source) & $\sim$100 & Not specified & Front facing dashcam clips; the released collection is about one hundred hours. \\

DR(eye)VE\linebreak \cite{palazzi2018predicting} & Not stated (urban, countryside, motorway routes) & 6.17$^{h}$ & Both & Roof mounted car perspective video; 74 sequences of 5 minutes each; recorded at daytime and at night (also includes weather variation). \\

HDD (Honda)~\cite{ramanishka2018toward} & USA (San Francisco Bay Area, CA) & 104 & Not specified & Instrumented vehicle dataset; this row refers only to the front facing stream (dashcam style viewpoint). \\

IDD\textendash CRS\cite{mishra2025idd} & India (unstructured traffic; 30 day collection) & 90 & Not specified & 5,400 untrimmed front view videos (1 min each) collected via dashcam; long tail critical road scenarios. \\

IDD\textendash X~\cite{parikh2024idd} & India (Hyderabad region; urban, rural, motorway) & 85 & Both & Dual view (front and rear) driving videos; captured across day and night and varied weather; front view is dashcam style. \\

KITTI (raw)~\cite{geiger2012we} & Germany (Karlsruhe) & 6 & Day & Vehicle mounted stereo camera; front facing viewpoint; raw driving sequences. \\

KITTI 360~\cite{liao2022kitti} & Germany (Karlsruhe) & N/A$^{i}$ & Not specified & Multi sensor platform; forward facing frames available; published primarily via distance and frame counts (for example 73.7 km) rather than total hours. \\

Mapillary Vistas~\cite{neuhold2017mapillary} & Six continents (not itemised) & N/A$^{r}$ & Not quantified & 25,000 street level still images from phones, action cameras and professional rigs, not video; included for its geographic scope. \\

nuPlan~\cite{caesar2021nuplan} & USA (Boston, MA; Pittsburgh, PA; Las Vegas, NV), Singapore & 1,200$^{j}$ & Not specified & Human driving dataset with multi sensor logs; forward camera stream available. \\

nuScenes~\cite{caesar2020nuscenes} & USA (Boston), Singapore & 5.56$^{k}$ & Both & 1,000 scenes $\times$ 20 s; 6 cameras (front included) + LiDAR/radar; includes night and rain. \\

ONCE~\cite{mao2021one} & Not fully enumerated (200 km$^2$ driving regions) & 144 & Both & 1M LiDAR scenes with 7M camera images; 7 cameras + LiDAR; diverse environments (day, night, sunny, rainy, urban, suburban). \\

OpenDV 2K (OpenDriveLab)~\cite{yang2024generalized} & $\ge$40 countries, $\ge$244 cities & 2,059 (1,747 YouTube) & Not quantified & Web mined front view driving videos paired with text; country and city counts are estimates from video titles; camera setup is described as uncalibrated. \\

Oxford RobotCar~\cite{maddern20171} & UK (Oxford) & N/A$^{l}$ & Both & Repeated route over $\sim$1,000 km; data collected across varied weather and lighting conditions; multi camera platform, using the forward view stream(s) as dashcam style video. \\

PandaSet~\cite{xiao2021pandaset} & USA (California: Silicon Valley, San Francisco) & 0.23$^{m}$ & Both & Multi sensor dataset with forward camera stream; 103 scenes of 8\,s each. \\

PhysicalAI-AV (NVIDIA)~\cite{nvidia_physicalai_av_2025} & 25 countries, 2,500+ cities & 1,727 & Both & 306,152 clips $\times$ 20 s; 7 RGB cameras including front wide and tele; access gated by NVIDIA AV dataset licence. \\

UDRIVE~\cite{eenink2014udrive} & Six European countries (Germany, France, Poland, the Netherlands, the UK, Spain) & $>$88,000$^{s}$ & Not specified & Naturalistic driving study of cars, trucks and powered two-wheelers with cameras and sensors in the vehicles; 19 months of data collection. \\

Waymo Open Dataset (Perception)~\cite{sun2020scalability} & USA (San Francisco, CA; Phoenix, AZ; Mountain View, CA) & 11.28$^{n}$ & Both & 2030 segments $\times$ 20 s; multi camera + LiDAR (forward view available); diverse conditions including night and varied weather. \\

ZOD (Zenseact Open Dataset)~\cite{alibeigi2023zenseact} & 14 European countries & 9.7 & Both & Overall dataset spans 14 countries; dashcam video includes 1,473 sequences of 20\,s (8.2\,h$^{o}$) and 29 drives totalling 1.5\,h. \\

\midrule
\textbf{CROWD (this work)} & 238 countries and territories (10,538 localities) & 36,204 & Both & Manually curated segments of ordinary urban driving from publicly available YouTube uploads (107,407 segments from 82,484 uploads); distributed as video identifiers, segment boundaries and detection and tracking pseudo-labels. \\

\end{xltabular}

\endgroup

Other related datasets emphasise repeated routes, long term variation, or challenging conditions, typically within a small number of operating areas. Oxford RobotCar provides repeated traversals of a fixed route under various conditions, including night, but the data are tied to a single city and are often summarised by distance or image counts rather than a simple hour based measure \cite{maddern20171}. Related long term resources such as 4Seasons~\cite{wenzel20204seasons} and Boreas~\cite{burnett2023boreas} target seasonal and weather variation on repeated routes, again prioritising controlled revisit structure over broad cross city coverage. Datasets such as CADC~\cite{pitropov2021canadian} focus on adverse conditions and winter driving, providing forward camera streams alongside other sensors, with a geographical scope restricted to a limited region.

A further set of benchmarks relies on instrumented vehicles and multi sensor acquisition to provide carefully captured urban driving sequences with extensive annotations. KITTI, KITTI-360, and A2D2 have been crucial to progress in autonomous driving research, but their geographic scope is limited to a small number of cities and collection campaigns \cite{geiger2012we,liao2022kitti,geyer2020a2d2}. ONCE provides large scale multi-sensor driving data reported by regions rather than a global set of cities \cite{mao2021one}. HDD and DR(eye)VE incorporate forward video alongside additional signals and support driver behaviour and attention related analyses, rather than functioning as globally distributed dashcam corpora \cite{ramanishka2018toward,palazzi2018predicting}. Comma2k19 provides a road facing a stream on a fixed commute route, which is useful for longitudinal studies on a single corridor, but does not represent a wide variation in urban life \cite{schafer2018commute}. Recent resources such as IDD-CRS and IDD-X provide important coverage of unstructured traffic in India, but remain geographically concentrated relative to a cross country design goal \cite{mishra2025idd,parikh2024idd}.

Web sourced collections can expand geographic breadth at scale by leveraging publicly hosted driving videos. OpenDV 2K reports broad coverage across at least 40 countries and at least 244 cities, illustrating the potential of web hosted material for diversity, while also inheriting heterogeneity in camera configuration and variability in metadata quality \cite{yang2024generalized}. Recent work on driving world models has highlighted the scalability of single view methods based on monocular ego video and the use of weakly curated Internet sourced driving footage for model adaptation \cite{rahimi2026mad}. PhysicalAI-AV provides large scale multi camera data in many locations, with access governed by a dataset licence \cite{nvidia_physicalai_av_2025}. Web scale datasets can therefore provide breadth, yet they are not necessarily curated to prioritise routine urban driving segments over atypical events or non driving intervals.

\begin{figure}[htbp]
  \includegraphics[width=\textwidth]{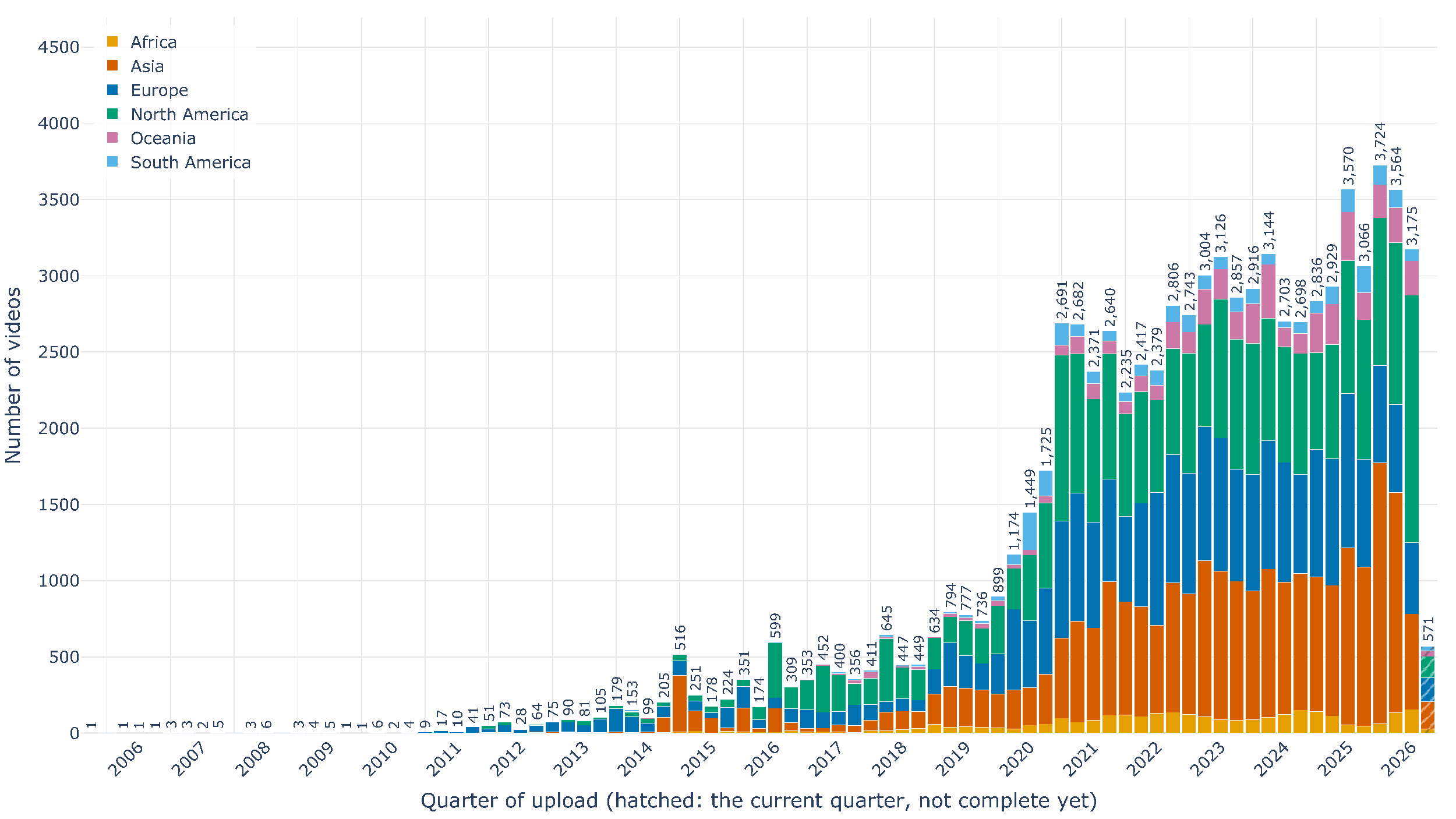}
  \caption{Quarterly platform upload date distribution of YouTube videos with at least one retained CROWD segment. Quarters are January to March, April to June, July to September, and October to December. This distribution reflects YouTube upload dates, not independently verified recording dates, and the date of driving may differ from the date of upload. The 2026 values cover only the incomplete period available at metadata freeze and should not be visually compared with complete calendar years. Metadata for this figure were frozen on 2 November 2026. Two unavailable or invalid upload-date entries were omitted from date parsing.}
  \label{fig:upload_month}
\end{figure}

Taken together, these design choices leave limited support for studies that require broad geographic coverage, routine urban driving, and longer continuous temporal context within a single resource. CROWD is intended to complement existing datasets by providing manually filtered, minute scale contiguous clips of ordinary urban driving from publicly available videos. In addition to the raw video segments, CROWD provides frame aligned object bounding boxes generated by a documented pipeline, together with structured data records and technical validation to support reproducible baselines.



\section{Methods}
\label{sec:methods}

\subsection{Video source and selection criteria}
A large volume of dashcam driving footage is available online, with YouTube (\url{https://www.youtube.com}) being a dominant hosting platform. Network traffic measurements report that YouTube accounts for about 16\% of downstream volume on fixed networks and about 21\% on mobile networks. \cite{sandvine2024gipr}. Much widely shared dashcam content is selected for entertainment value and often concentrates on atypical or high conflict events (for example, crashes, near misses, and confrontations), exemplified by curated sharing channels such as the Telegram group BadShofer (\url{https://t.me/s/badshofer}), 40,716 subscribers, accessed 31 March 2026). Such clips are typically short, incident focused, and frequently include changes in viewpoint or editing, making them poorly suited for analyses of routine urban exposure.

In contrast, a separate genre on YouTube consists of long, continuous recordings captured with relatively stable, forward facing equipment (often longer than 40 minutes per upload; for example: \url{https://www.youtube.com/@jutah}, approximately 834,000 subscribers and 894 videos, accessed 31 March 2026). These videos are commonly consumed for their relaxing, monotonous visual and auditory characteristics and can trigger Autonomous Sensory Meridian Response (ASMR)~\cite{lohaus2023effects, asmr_vox_2015, barratt2017sensory} or serve as ambient background viewing~\cite{han2020youtubers}. Previous work on ASMR on YouTube identified driving themed content as a recurring theme within ASMR videos \cite{alam2026ASMR}. We targeted the latter category to curate footage representative of routine urban driving.


\subsection{Search strategy, screening, segmentation, and manual labels}

Candidate videos were identified manually by the authors through YouTube search using phrases such as \textit{dashcam driving in [city]}, \textit{driving video in [city]}, \textit{dashcam videos in [country]}, and \textit{dashcam driving in [country]}, where the bracketed terms were replaced by the locality or country names targeted during curation. The input records were publicly viewable YouTube uploads; no videos were obtained from private repositories, paid datasets, organisational collaborations, or non-public data sharing agreements. To allow others to locate the same source records where they remain available, the deposited \texttt{mapping.csv} and \texttt{mapping\_metadata.csv} files include the YouTube upload identifier for each retained video. Each source upload can be resolved using the standard YouTube watch URL pattern \url{https://www.youtube.com/watch?v=VIDEO_ID}, replacing \texttt{VIDEO\_ID} with the corresponding identifier in the release. Throughout this paper, we use \textit{locality} to denote the named inhabited place associated with a record, including hamlets, villages, towns and cities~\cite{UNStatsLocality2017}. All identification, screening, and segmentation decisions were performed by the authors using a shared protocol (rather than through large scale crowd sourced annotation), to support consistent application of the inclusion and exclusion criteria. We restricted candidates to videos that are publicly viewable on YouTube and accessed through features of the service in accordance with the YouTube Terms of Service and the uploader selected YouTube licence type, noting that YouTube supports the Standard YouTube licence by default and the Creative Commons Attribution licence where specified (\url{https://support.google.com/youtube/answer/9783148?hl=en&sjid=15816069292466875886-EU}). Then each candidate upload was manually selected and only continuous driving portions that met the eligibility criteria below were retained. Screening consisted of a direct visual review of the footage to verify compliance and determine segment boundaries, as follows:

\begin{itemize}
    \item \textbf{Source upload and segment duration:} candidate source uploads were required to be at least five minutes long. This threshold applies to the original YouTube upload, not to the retained segment length. Retained CROWD segments are variable duration compliant intervals, not fixed length five minute windows. Each segment preserves the full continuous portion of an upload that met the eligibility criteria.

    \item \textbf{Predominantly urban:} segments primarily depict streets and junctions within urban areas. This includes ordinary city streets, as well as larger urban roads, such as inner city arterial roads, boulevards, ring roads, and wide multilane roads, when they remain embedded in the urban environment and show urban traffic context such as junctions, pedestrian crossings, sidewalks, roadside buildings, bus stops, parked vehicles, pedestrians, cyclists, or frequent local access. In large car oriented metropolitan areas, some retained footage may also include major urban road corridors within the city, such as parts of Interstate 10 in Los Angeles (CA), where the surrounding scene remains visibly urban. Inclusion is based on the observed roadside context rather than the road name or classification alone. Short intervals are allowed through urban parks or green spaces within a locality. Segments dominated by limited access motorways or freeways, rural roads, large parking areas, or other clearly nonurban contexts are excluded, especially when they lack observable urban roadside activity or vulnerable road user infrastructure.

    \item \textbf{Routine conditions:} segments dominated by atypical events or unusual disruptions are excluded, for example, collisions and their aftermath, near misses, confrontations, police stops, emergency response scenes, parades, protests, festivals, or other organised events that materially alter the driving context.
    

    \item \textbf{Continuous unedited dashcam view:} segments containing edited discontinuities or non-dashcam intervals are excluded. Edited discontinuities are defined as explicit edits that remove intervening time or change context, including hard cuts or jump cuts, cross fades or other transitions, inserted title cards or chapter separators that skip to a different time or location, time lapse or accelerated playback, and montage or compilation structures.

    \item \textbf{Normal playback speed and no advertisements:} segments must play at normal ($1\times$) speed, so footage that is sped up or slowed down is excluded. Advertisements inserted into the video by the uploader or the platform are not footage of the road and are excluded; the footage on either side of an advertisement is retained as separate segments.

    \item \textbf{On-screen text and overlays:} videos sometimes contain captions, watermarks, channel branding, timestamps, or other labels. These are permitted when they do not substantially occlude the driving scene. Candidates with persistent overlays that occupy a substantial portion of the frame are excluded.

    \item \textbf{Stationary non-traffic stops:} intervals where the vehicle is stationary for reasons unrelated to normal traffic flow are excluded, including prolonged stops in car parks, petrol stations, service areas, drive through queues, and similar situations. This criterion is applied on the basis of the scene context and the duration of the stationary interval. 
\end{itemize}

\begin{table}[!htbp]
\centering
\caption{Distribution of retained CROWD segments and total retained duration by continent, with counts of represented countries and territories. \emph{Segment share} is computed over all retained segment records; \emph{duration share} is computed over total retained hours. Continents follow the UN Statistics Division (UNSD) M49 geographic regions convention~\cite{UNSD_M49}.}
\resizebox{\linewidth}{!}{%
\begin{tabular}{lrrrrr}
\toprule
Continent & Countries & Segment records & Segment share (\%) & Duration share (\%) & Duration (h) \\
\midrule
Europe        & 53 & 29,743 & 27.69 & 26.29 & 9,517.19 \\
Asia          & 52 & 32,534 & 30.29 & 28.33 & 10,255.11 \\
North America & 37 & 30,810 & 28.69 & 33.46 & 12,113.40 \\
Africa        & 59 & 4,259 & 3.97 & 2.97 & 1,074.87 \\
Oceania       & 22 & 6,112 & 5.69 & 5.30 & 1,918.87 \\
South America & 18 & 3,949 & 3.68 & 3.66 & 1,324.83 \\
\midrule
Total         & 238 & 107,407 & 100.00 & 100.00 & 36,204.26 \\
\bottomrule
\end{tabular}%
}

\label{tab:continent-segments}

\par\vspace{3pt}
\noindent\footnotesize\justifying Note: Countries and territories are counted within each continent, so a country or territory whose localities are assigned to more than one continent appears in more than one row; the total counts each country or territory once. The segment and duration totals in Table~\ref{tab:continent-segments} are computed by summing over mapping rows within each continent. In this release, no upload is associated with localities in more than one continent. For counts of unique uploads by continent, each upload is nevertheless assigned to a single continent by a fixed rule: we first take the most frequent continent among its mapped rows; if there is a tie, we assign the upload to the continent with the greatest total mapped duration (in seconds) for that upload.
\end{table}

When a source upload contained compliant driving footage, each continuous compliant interval was retained as a variable duration segment, with its start and end times recorded in seconds relative to the beginning of the original upload. The segments were defined by compliance boundaries rather than by a fixed target duration. If excluded content occurred within an otherwise eligible upload, such as a crash scene, non dashcam interval, edited transition, or prolonged stationary non traffic stop, the affected portion was omitted, and the remaining compliant intervals were retained as separate segments where applicable. Ordinary traffic related stopping could remain within a segment. Retained intervals from the same upload were required to be non overlapping, and candidate overlaps identified during curation or release validation were resolved before release. Across the dataset, the 107,407 segment records span 36,204.26 h of processed footage, corresponding to a mean processed segment duration of 20.22 min (median 14.52 min; 10th to 90th percentile 3.53 to 43.85 min; shortest segment 5 s). A segment can be much longer than the five-minute minimum for an upload, because it keeps the whole continuous compliant interval; the longest is 602.6 min. Searches and screening were conducted from 17 December 2023 to 2 November 2026.

In addition to this manual search, candidate segments were also proposed semi-automatically for the channels of uploaders who publish long continuous driving videos. For each video of a channel, a low-resolution copy (160$\times$90 pixels, analysed at two frames per second) was used to propose segment boundaries, a locality, a time-of-day label and a platform category. Videos were excluded in advance from their platform metadata when they were shorter than five minutes, live streams, portrait videos, walking tours, time-lapses, compilations, crash videos, highway drives or road trips. Within a retained video, stationary stretches at the start and end were trimmed and stops longer than three minutes were cut out; hard cuts, blank frames and title cards ended a segment; stretches with the camera not facing forward, shaking, or on rural roads or highways were left out; and proposals shorter than 30 s were dropped. These proposals were never added to the dataset automatically. A curator reviewed every proposed video on a web page that shows the player, the proposed segments and the reason for each cut, and could change, split, reject or approve each segment. Only approved segments were added to \texttt{mapping.csv}, with the same overlap checks and label sets as manually entered segments.

\paragraph{Country and locality concentration.}
To characterise the distribution of retained footage within the geographic coverage reported above, we quantified the retained processing duration by country and by location, using the same endpoint adjustment applied to the duration totals. Table~\ref{tab:country-locality-concentration}(a) reports the ten largest contributing countries by the processing duration retained; together, these countries account for 60.47\% of the retained hours. Table~\ref{tab:country-locality-concentration}(b) reports the ten largest contributing localities by retained processed duration. At the locality level, the top 1\% of localities (106 localities) contribute 68.66\% of retained hours, the top 5\% (527 localities) contribute 84.78\%, and the top 10\% (1,054 localities) contribute 89.35\%. In contrast, 5,491 localities (52.11\%) are represented by exactly one segment, and 6,266 localities (59.46\%) are represented by exactly one unique upload.
\begin{table}[!htbp]
\centering
\caption{Largest contributing countries and localities by retained processed duration. Shares are computed over total retained processed hours.}
\label{tab:country-locality-concentration}
\footnotesize
\textbf{(a) Countries}\\[2pt]
\resizebox{\linewidth}{!}{%
\begin{tabular}{rlrrrrr}
\toprule
Rank & Country & Duration (h) & Share (\%) & Segment records & Unique uploads & Localities \\
\midrule
1 & United States & 10,511.09 & 29.03 & 25,368 & 20,565 & 2,359 \\
2 & Australia & 1,728.96 & 4.78 & 5,133 & 3,982 & 213 \\
3 & United Kingdom & 1,703.93 & 4.71 & 3,606 & 3,151 & 379 \\
4 & Russia & 1,522.87 & 4.21 & 4,070 & 3,201 & 212 \\
5 & Philippines & 1,349.98 & 3.73 & 4,388 & 3,654 & 414 \\
6 & Canada & 1,273.68 & 3.52 & 4,114 & 3,625 & 239 \\
7 & Vietnam & 1,196.79 & 3.31 & 3,852 & 3,054 & 79 \\
8 & Indonesia & 1,033.56 & 2.85 & 3,530 & 3,169 & 177 \\
9 & South Korea & 822.83 & 2.27 & 1,323 & 838 & 84 \\
10 & France & 750.39 & 2.07 & 2,900 & 1,863 & 369 \\
\bottomrule
\end{tabular}%
}

\vspace{3mm}
\textbf{(b) Localities}\\[2pt]
\resizebox{\linewidth}{!}{%
\begin{tabular}{rllrrrr}
\toprule
Rank & Locality & Country & Duration (h) & Share (\%) & Segment records & Unique uploads \\
\midrule
1 & New York, NY & United States & 2,280.45 & 6.30 & 4,820 & 4,179 \\
2 & Chicago, IL & United States & 1,713.92 & 4.73 & 2,362 & 1,896 \\
3 & London & United Kingdom & 1,363.36 & 3.77 & 2,275 & 2,038 \\
4 & Los Angeles, CA & United States & 1,164.78 & 3.22 & 2,840 & 2,047 \\
5 & Hanoi & Vietnam & 875.96 & 2.42 & 2,697 & 2,151 \\
6 & Manila & Philippines & 793.55 & 2.19 & 2,694 & 2,179 \\
7 & Sydney, NSW & Australia & 682.04 & 1.88 & 1,580 & 1,432 \\
8 & Saint Petersburg & Russia & 566.02 & 1.56 & 787 & 689 \\
9 & Seoul & South Korea & 543.04 & 1.50 & 662 & 294 \\
10 & Jakarta & Indonesia & 474.99 & 1.31 & 1,774 & 1,674 \\
\bottomrule
\end{tabular}%
}
\end{table}



To standardise ingestion and minimise duplicate locality entries, we used a structured curation workflow implemented as a web based form. For each candidate upload, the curator entered a locality name, an optional state or region, the country, and the YouTube URL. The locality names were normalised using a canonical \textit{locality} field and a companion \textit{locality aliases} field, which records alternative spellings as well as historical or local names for the same locality (for example, \textit{Copenhagen}, \textit{Kobenhavn}, \textit{$K\o benhavn$}). Submitted names were matched against both canonical names and recorded aliases, with country and, when provided, state or region used to disambiguate homonymous localities. When a matching locality record already existed, new uploads were appended to that record; otherwise, a new locality record was created. We record aliases primarily in Latin script, and alternative names written only in non Latin scripts are not systematically included in the \textit{locality aliases} field.

For each submitted URL, the system resolved the YouTube upload identifier and retrieved basic platform metadata needed for provenance, including the platform upload date where available. The dataset does not include a recording date field because the date of driving could not be independently verified consistently at scale from the source records. Dates mentioned in video titles, descriptions, chapters, or visual overlays were therefore not treated as recording dates. Accordingly, no recording date, recording date provenance, or recording date uncertainty fields are included in this release.

The curator then specified segment start and end times in seconds relative to the beginning of the upload and assigned manual labels for time of day and recording platform type. At submission, the resolved YouTube upload identifier was checked against all existing mapping records. When the identifier was already present, the interface displayed its existing locality associations and retained segment intervals. A proposed segment was rejected if its interval overlapped any segment already recorded for the same upload, either within the current locality record or elsewhere in the mapping; intervals that shared only an endpoint were permitted. This prevented the same footage interval from being retained more than once while allowing distinct, non-overlapping portions of one upload to be associated with different localities. Additional validation checks enforced that the end time exceeded the start time and that categorical fields took values from the permitted label sets.


\begin{table}[!htbp]
\centering
\caption{Upload-level time-of-day composition within each continent. An upload is \emph{day only} if all retained segments are labelled day, \emph{night only} if all are labelled night, and \emph{both} if it includes at least one of each. Uploads are assigned a canonical continent (mode over linked segments) so that per-continent totals sum to the global unique-upload total. Percentages are computed within continent.}
\begin{tabular}{lrrrr}
\toprule
Continent & Day & Night & Both day \& night & Unique uploads \\
\midrule
Europe        & 19,833 (88.45\%) & 1,978 (8.82\%) & 612 (2.73\%) & 22,423 \\
Asia          & 19,996 (81.60\%) & 3,604 (14.71\%) & 905 (3.69\%) & 24,505 \\
North America & 21,552 (86.05\%) & 2,515 (10.04\%) & 980 (3.91\%) & 25,047 \\
Africa        & 2,842 (93.70\%) & 140 (4.62\%) & 51 (1.68\%) & 3,033 \\
Oceania       & 4,068 (89.54\%) & 322 (7.09\%) & 153 (3.37\%) & 4,543 \\
South America & 2,399 (81.79\%) & 420 (14.32\%) & 114 (3.89\%) & 2,933 \\
\midrule
Total         & 70,690 (85.70\%) & 8,979 (10.89\%) & 2,815 (3.41\%) & 82,484 \\
\bottomrule
\end{tabular}
\label{tab:continent-daynight-uploads}
\end{table}

Each retained segment was assigned a binary contextual time-of-day label by visual inspection. A segment was labelled night when street lamps or comparable artificial road lighting were visibly active and consistently present in the driving scene; otherwise it was labelled day. If retained footage transitioned between daylight and night under this rule, it was split at the transition so that each segment had one label. The field is available for every retained segment and is intended for broad dataset stratification; it does not distinguish dawn, dusk, or twilight, nor does it represent astronomical time or measured illumination.

The recording platform label is also complete under the original curation protocol and describes the apparent recording platform or source asserted platform category available during curation. The represented classes are \textit{Automated car}, \textit{Bicycle}, \textit{Bus}, \textit{Car}, \textit{Electric scooter}, \textit{Emergency vehicle}, \textit{Truck}, \textit{Two-wheeler}, and \textit{Monowheel/unicycle}. When platform type was difficult to infer from the forward facing view, the authors used additional contextual cues, including the vehicle silhouette, shadow, or reflection when visible, for example, in windows or in the bodywork of nearby vehicles, apparent camera height and motion, and the width and vertical clearance of the path being followed, for example, the clearance and lane use typically required by a car compared with a bicycle. The \textit{Automated car} label was assigned only when source provided title, description, or chapter information indicated automated driving; automation status was not inferred from camera height, shadows, or other visual cues alone. These labels should be treated as apparent or source asserted categories, not as independently verified automated driving status.

To improve the consistency of manual labels and segment boundary definitions, we incorporated an independent audit step during curation. One author periodically reviewed a random sample of segments added in the preceding two weeks by the other authors. For each sampled segment, the auditor re-watched the footage and verified segment start and end times, the time-of-day label, and recording platform type. Any discrepancies were corrected in the curation database, and ambiguous cases were resolved by discussion among the authors. The sizes of these early audit samples and the numbers of corrections were not logged, so they cannot be reported. To quantify the reliability of the boundaries and labels after the fact, we drew a stratified random sample of segments for a formal audit, which is reported in the Technical validation section. The released time-of-day and platform fields should nevertheless be used as coarse contextual metadata rather than benchmark-grade ground truth.


\begin{table}[!htbp]
\centering
\caption{Global unique-upload counts by apparent recording platform category. Percentages are computed relative to the global unique-upload total ($n=82,484$). An upload that carries different labels in different locality records (13 uploads) is counted under its most frequent label.}
\begin{tabular}{lrr}
\toprule
Vehicle type & Videos & Share (\%) \\
\midrule
Automated car      & 24 & 0.03 \\
Bicycle            & 1,475 & 1.79 \\
Bus                & 1,104 & 1.34 \\
Car                & 77,265 & 93.67 \\
Electric scooter   & 88 & 0.11 \\
Emergency vehicle  & 314 & 0.38 \\
Monowheel/unicycle & 38 & 0.05 \\
Truck              & 751 & 0.91 \\
Two-wheeler        & 1,425 & 1.73 \\
\bottomrule
\end{tabular}
\label{tab:vehicle-tod-global}
\par\vspace{3pt}
\noindent\footnotesize\justifying Note: The platform categories are complete under the original curation protocol and are intended for broad filtering and dataset characterisation. They should not be interpreted as independently verified platform identity. In particular, automated categories indicate apparent or source asserted status available during curation and should not be treated as verified automated-driving ground truth.\normalsize
\end{table}



\subsection{Sampling frame and selection flow}
CROWD is a non-probability convenience sample. Videos were found by manual search and by channel, so only footage that uploaders chose to publish, and that the platform made findable, could enter. There is no sampling frame of driving, and the probability that a place or a road is included is unknown and differs between countries, for example through the number of uploaders, the language of titles and the ranking of the search. Geographic coverage therefore describes where suitable footage was found. It is not a measure of how much driving takes place there, and it should not be read as representative of driving exposure within or across countries. \autoref{fig:flow} summarises the selection flow. The number of uploads that were screened and rejected during manual search was not recorded, so inclusion rates cannot be reported for the stages before screening. Guidance for analyses that compare countries is given in the Usage Notes.

\begin{figure}[htbp]
\centering
\begin{tikzpicture}[node distance=6mm, box/.style={draw, rounded corners, align=left, text width=0.86\textwidth, inner sep=4pt, font=\small}, arr/.style={-{Stealth[length=2.5mm]}, thick}]
\node[box] (a) {\textbf{1. Identification.} Manual YouTube searches by locality and country, and channel-level proposals for uploaders of long continuous driving videos. The number of candidates was not recorded.};
\node[box, below=of a] (b) {\textbf{2. Eligibility.} Publicly viewable upload of at least five minutes, forward facing, and of a continuous driving recording.};
\node[box, below=of b] (c) {\textbf{3. Screening and segmentation.} Each continuous stretch that meets the criteria (urban, routine, unedited, normal speed, no advertisements, no prolonged stops) is retained as a segment with start and end times.};
\node[box, below=of c] (d) {\textbf{4. Labels and registration.} Locality, time of day and platform type; overlap checks against all existing segments of the upload.};
\node[box, below=of d] (e) {\textbf{5. Second screen.} Scene-change screen of the retained segments (52,838 candidate cuts), reviewed by hand; confirmed cuts split a segment.};
\node[box, below=of e] (f) {\textbf{6. Release.} 82,484 uploads, 107,407 segments in 10,538 localities, with detections and tracks.};
\foreach \from/\to in {a/b,b/c,c/d,d/e,e/f} \draw[arr] (\from) -- (\to);
\end{tikzpicture}
\caption{Selection flow of CROWD. Numbers are given where they were recorded.}
\label{fig:flow}
\end{figure}

\subsection{Video retrieval and processing}

Videos were downloaded from YouTube as MP4 files; a video that had already been downloaded was reused from a local file server instead of being downloaded again. The pipeline first attempted retrieval with \textit{pytubefix}~(\url{https://pytubefix.readthedocs.io/en/latest/}) and if that failed, retried with \textit{yt\_dlp} \linebreak (\url{https://pypi.org/project/yt-dlp/}). In both cases, a single stream was selected by resolution. The downloader first searched for an exact match among the preferred resolutions of 720p, 480p, 360p, and 144p. When no exact match was available, it selected the highest available MP4 stream with frame height at most 720 pixels, preferring a progressive stream when multiple streams shared the same height. If no MP4 stream at or below 720 pixels was available, it selected the lowest available MP4 stream above 720 pixels, again preferring a progressive stream when possible. In the \textit{yt\_dlp} fallback, only non HLS MP4 video formats were considered. After download, the frame rate was obtained from the saved video file using OpenCV (\url{https://opencv.org}) and rounded to the nearest integer. This rounded FPS value was then used in the output file names and in subsequent segment processing and tracking. The rounding has a small effect on timing: 29.95\% of the source videos are 29.97 fps but were processed at 30 fps, so a frame index converted to seconds with the rounded rate is too small by about 0.1\%, or about 1.2 s over a segment of the mean duration. Users who need exact timing should read the frame rate of the video itself.

Each retained upload was trimmed into the manually curated segments. The released \verb|start_time| and \verb|end_time| fields store the original manually selected segment boundaries in seconds relative to the beginning of the YouTube upload. To avoid boundary artefacts during trimming, the temporary videos used for detection and tracking were generated with the processing endpoint $t_{\mathrm{end,processed}}=t_{\mathrm{end}}-1~\mathrm{s}$, with a safeguard that leaves the endpoint unchanged if the adjusted interval would be non-positive. Duration totals reported here are computed as $t_{\mathrm{end,processed}}-t_{\mathrm{start}}$. Across all 107,407 segments, this gives 130,335,349 s (36,204.26 h) of processed footage after applying the endpoint adjustment. For a segment processed at rounded frame rate $f_i$, the expected frame-count reduction is one second of footage, that is $f_i$ frames at the rounded frame rate. Frame indices in the released per-segment CSV files are segment-local: frame 1 corresponds to the first processed frame at the segment start, and frame $k$ corresponds approximately to $t_{\mathrm{start}}+(k-1)/f_i$ seconds in the original upload.


All retained segments were processed with the You Only Look Once (YOLO) object detector \cite{redmon2016you} using Ultralytics YOLO11x weights (\verb|yolo11x.pt|) trained on MS COCO \cite{lin2014microsoft}. The model weights have SHA256 checksum \nolinkurl{7bc158aa95c0ebfdd87f70f01653c1131b93e92522dbe15c228bcd742e773a24}. For each processed frame, the detector produced bounding boxes, MS COCO class labels, and confidence scores. The pinned software environment and effective inference settings used to generate the released detection and tracking pseudo-labels are reported in \autoref{tab:detection_environment}. The machine-readable configuration files are documented with the source code (see the \hyperref[sec:code_avail]{Code Availability section}).

\begin{table}[!htbp]
\centering
\caption{Schema of \texttt{mapping.csv}. List valued fields are stored as text using square bracket notation, for example \texttt{[a,b,c]}. Indices align with \texttt{videos}. Nested lists in \texttt{time\_of\_day}, \texttt{start\_time}, and \texttt{end\_time} represent multiple retained segments within one upload.}
\small
\setlength{\tabcolsep}{3pt}
\renewcommand{\arraystretch}{1.05}

\begin{tabular}{l l l p{0.62\textwidth}}
\toprule
Field & Type & Units & Notes \\
\midrule
\texttt{id} & int & -- & Locality record identifier, unique per row, $>0$. \\
\texttt{locality} & string & -- & Canonical locality name. \\
\texttt{locality\_aka} & string & -- & Locality aliases stored as a text encoded list (for example \texttt{["Kobenhavn","K\o benhavn"]}). \\
\texttt{state} & string & -- & Optional state or region, may be empty. \\
\texttt{country} & string & -- & Country or territory name. \\
\texttt{iso3} & string & -- & ISO3 code. \\
\texttt{continent} & string & -- & Geographic continent from coordinates. \\
\texttt{lat} & float & degrees & Latitude in $[-90,90]$. \\
\texttt{lon} & float & degrees & Longitude in $[-180,180]$. \\
\texttt{gmp} & float & -- & Gross metropolitan product when available, 0.0 if unavailable. \\
\texttt{population\_locality} & int & persons & Locality population estimate retrieved through an API where available; otherwise entered manually after online lookup, $\ge 0$. \\
\texttt{population\_country} & int & persons & Country or territory population retrieved through an API where available; otherwise entered manually after online lookup, $\ge 0$. \\
\texttt{traffic\_mortality} & float & -- & Road traffic mortality rate retrieved programmatically where available; otherwise entered manually after online lookup. \\
\texttt{literacy\_rate} & float & \% & Literacy rate retrieved programmatically where available; otherwise entered manually after online lookup, in $[0,100]$. \\
\texttt{avg\_height} & float & cm & National average height, $\ge 0$. \\
\texttt{med\_age} & float & years & Median age, $\ge 0$. \\
\texttt{gini} & float & -- & Income inequality indicator retrieved programmatically where available; otherwise entered manually after online lookup, typically in $[0,100]$. \\
\texttt{traffic\_index} & float & \% & Percentage by which travel on the road nearest to the locality was slower than free flow at the time of a TomTom query, $\ge 0$; empty where no reading is available. In older entries, a failed request may be stored as 0. \\
\texttt{videos} & string & -- & List of YouTube video IDs stored as text, for example \texttt{[QZPqluJA0OY,...]}. \\
\texttt{time\_of\_day} & string & -- & Nested lists stored as text. Outer indices align with \texttt{videos}. Inner lists give one binary contextual label per segment, where 0 means day and 1 means night. \\
\texttt{start\_time} & string & seconds & Nested lists stored as text. Segment start times in seconds from upload start. Outer indices align with \texttt{videos}. \\
\texttt{end\_time} & string & seconds & Nested lists stored as text. Segment end times in seconds from upload start. Outer indices align with \texttt{videos}. \\
\texttt{vehicle\_type} & string & -- & List stored as text with one apparent recording platform or source asserted platform label per video, aligned with \texttt{videos}; automated status is not independently verified. \\
\texttt{upload\_date} & string & -- & List stored as text with one platform upload date per video where retrievable, aligned with \texttt{videos}, stored as \texttt{DDMMYYYY} without a leading zero for days 1 to 9 (for example 7052021); this is not a recording date. \\
\texttt{channel} & string & -- & List of platform-assigned channel identifiers aligned with \texttt{videos} where retrievable; these are not channel names, handles, or custom URLs. \\
\bottomrule
\end{tabular}

\label{tab:mapping-schema}
\par\vspace{3pt}
\noindent\footnotesize\justifying Note: Complete platform metadata (title, upload date, channel and view count were all retrieved) was not available for 3,267 of the 82,484 unique uploads (3.96\%). For 2,762 of these, nothing was returned (no title, channel or view count) when metadata retrieval was attempted, consistent with uploads having been removed, made private, or otherwise becoming unavailable; the specific reason could not be determined consistently. The remaining 505 have a title, channel and view count but no upload date.
\end{table}
Missing values are left empty and remain aligned by index with \texttt{videos}. This can affect \texttt{upload\_date}, \texttt{channel}, and other retained platform metadata fields, but not the retained segment boundaries or manually curated labels. The \texttt{channel} field stores a platform-assigned channel identifier, not a channel name, handle, or custom URL.
\normalsize


Detections were linked across frames using the BoT-SORT tracker \cite{aharon2022botsort} with the custom YAML configuration reported in \autoref{tab:tracking_hyperparams}. The configuration file has SHA256 checksum \nolinkurl{b0abba37f9e4a80451bf50afbe39bee9b54d48778e0ac988741d1899db9a41d8}. It specifies \verb|tracker_type=botsort|, enables re-identification (\verb|with_reid=true|), and uses \verb|sparseOptFlow| for global motion compensation. The YAML file contains an initial \verb|track_buffer| value of 60 at the 30 fps reference rate. Before each segment is processed, this value is overwritten at runtime by setting \verb|track_buffer = 2 * FPS|, using the rounded frame rate of that segment. The effective buffer is therefore frame-rate dependent and corresponds to two seconds of lost-track retention at the rounded frame rate of every segment. The ReID model setting was \verb|model=auto|; no separate ReID weight file was specified, and resolution of this setting followed the behaviour of the pinned Ultralytics 8.4.46 implementation. Tracking was reinitialised at each segment boundary; therefore, track identifiers are unique only within a segment.

The resulting bounding boxes and track identifiers were generated using pretrained Ultralytics YOLO11x weights trained on MS COCO together with the specified BoT-SORT configuration. The detector was not fine tuned on this dataset, and its object detection and tracking accuracy was evaluated only on a stratified sample of the CROWD videos (see Technical validation). Performance reported for the pretrained model on its original benchmark data was therefore not assumed to transfer directly to this dataset. The outputs were not manually corrected and are released as detection and tracking pseudo labels rather than object level ground truth. Their accuracy may vary by object class, country, locality, lighting, camera quality, traffic density, occlusion, and weather. Downstream studies requiring ground truth annotations should conduct task specific validation or manual review.

\begin{table}[!htbp]
\centering
\caption{Schema of \texttt{mapping\_metadata.csv}. This table has one row per YouTube upload for which metadata retrieval was attempted; fields are empty where retrieval failed. The \texttt{video} field links to the \texttt{videos} field of \texttt{mapping.csv}. The \texttt{channel} field stores a platform-assigned channel identifier, not a channel name or handle. The \texttt{title}, \texttt{description} and \texttt{chapters} fields hold text written by the uploaders and retrieved from the platform. They are not covered by the licence of the dataset and are not copied to \texttt{crowd\_index.jsonl}.}
\small
\setlength{\tabcolsep}{3pt}
\renewcommand{\arraystretch}{1.05}

\begin{tabular}{l l l p{0.62\textwidth}}
\toprule
Field & Type & Units & Notes \\
\midrule
\texttt{id} & int & -- & Video record identifier, $>0$. \\
\texttt{video} & string & -- & YouTube video identifier used for linking, for example \texttt{ID\_q1RC\_dSo}. \\
\texttt{title} & string & -- & Title of the upload as retrieved from the platform; third-party text written by the uploader. \\
\texttt{upload\_date} & int & -- & Platform upload date stored as \texttt{DDMMYYYY} without a leading zero for days 1 to 9, may be empty if unavailable; this is not a recording date. \\
\texttt{channel} & string & -- & Platform-assigned channel identifier; not a channel name, handle, or custom URL. \\
\texttt{views} & int & count & View count at metadata retrieval time, $\ge 0$ when available. \\
\texttt{description} & string & -- & Description of the upload as retrieved from the platform; third-party text written by the uploader. \\
\texttt{chapters} & string & -- & Chapter titles and timestamps of the upload as retrieved from the platform, stored as text; third-party text written by the uploader. \\
\texttt{segments} & int & count & Number of locality records in which this video appears. It is a non-authoritative summary; the retained segments are defined by \texttt{mapping.csv} and \texttt{crowd\_index.jsonl}. \\
\texttt{date\_updated} & int & -- & Date this metadata record was last updated, stored as \texttt{DDMMYYYY}. \\
\bottomrule
\end{tabular}

\label{tab:mapping-metadata-schema}
\end{table}


\subsection{Contextual indicators and metadata}
To support cross-locality comparisons, each locality record was enriched with contextual indicators retrieved at data entry time using the submitted locality, optional state/region, and country. Geographic coordinates (latitude and longitude) were recorded to support visualisation and location-linked queries. When available, coordinates were taken directly from video metadata; when coordinates could not be retrieved, they were obtained via manual online lookup using a standardised location query (\textit{locality, state/region (if available), country}) and entered during curation.

The resulting locality record includes population, road traffic mortality, income inequality, gross metropolitan product (when available), literacy rate, and a traffic index, along with geographic coordinates and a continent assignment based solely on geographic location. Locality population, country or territory population, road traffic mortality, literacy rate, Gini, and the traffic index were first retrieved programmatically through the relevant API or source service used by the curation pipeline. The locality query used the locality, optional state or region, and country. When a service did not return a usable value for any of these indicators, the curator searched publicly available internet sources and entered the value manually. These indicators should therefore be treated as contextual estimates because their sources and reference years may vary. National level indicators are assigned using a sovereign country mapping, which can differ from the local territory name in the \texttt{country} field. For example, Cayenne is geographically in South America, but national indicators are taken from France; similarly, Spanish enclaves in North Africa are geographically assigned to Africa, while national indicators are taken from Spain. Source services used in the curation pipeline include REST Countries (\url{https://restcountries.com/}) for country attributes (for example, population, ISO codes, and Gini), the World Bank indicators for traffic mortality (\url{https://data.worldbank.org/indicator/SH.STA.TRAF.P5}) and literacy (\url{https://data.worldbank.org/indicator/SE.ADT.LITR.ZS}), and the TomTom Traffic API for a location-based traffic index (\url{https://developer.tomtom.com/}), defined as the percentage by which travel on the road nearest to the locality was slower than free flow when the service was queried. Locality populations and coordinates were obtained from GeoNames (\url{https://www.geonames.org/}). Auxiliary tables were used to provide summary demographic statistics such as median age and average height. The traffic index is a snapshot that depends on the time of the query, and it is empty where TomTom has no coverage (for example Ukraine, Russia, Belarus and China). Earlier entries came from a different source (Numbeo), at times as a country-level value; in October 2026, 1,782 such values were replaced by local TomTom readings and 2,062 were cleared where TomTom has no coverage. Of the zeros that remain, it is not known which are genuine readings of free-flowing traffic and which are failed requests.

For each retained segment, the public release provides the source upload identifier and address, segment boundaries, manual labels, contextual locality metadata, and derived detection and tracking outputs. The platform-assigned channel identifier is retained solely to group uploads originating from the same source, quantify source concentration, detect duplicate source associations, and support provenance audits when individual uploads become unavailable. It is not accompanied by the channel name, handle, custom URL, profile information, or inferred uploader attributes. The title, description and chapter text of each upload are kept in \texttt{mapping\_metadata.csv} as retrieved from the platform, as part of the source metadata of the upload. This text was written by the uploaders, is not covered by the licence of the dataset, and is not copied to \texttt{crowd\_index.jsonl}. To minimise the redistribution of potentially identifying or copyrighted source material, the release excludes source videos, frames, audio, thumbnails, licence-plate transcriptions, identity labels, biometric descriptors, and other personal-information annotations.


\begin{table}[!htbp]
\centering
\caption{Top-level schema of \texttt{crowd\_index.jsonl}. Each line is one valid JSON object representing one retained segment. Nested objects use standard JSON types rather than Python-style list strings.}
\label{tab:jsonl-schema}
\small
\setlength{\tabcolsep}{4pt}
\renewcommand{\arraystretch}{1.05}
\begin{tabularx}{\textwidth}{l l l X}
\toprule
Field & Type & Units & Notes \\
\midrule
\texttt{video\_id} & string & -- & YouTube upload identifier. \\
\texttt{segment\_id} & string & -- & Unique identifier derived from the upload ID and segment boundaries. \\
\texttt{youtube\_url} & string & -- & Resolvable source watch address. \\
\texttt{start\_time\_s} & number & seconds & Retained segment start relative to the source upload. \\
\texttt{end\_time\_s} & number & seconds & Retained segment end relative to the source upload. \\
\texttt{upload\_date} & int/null & -- & Platform upload date where available, as \texttt{DDMMYYYY} without a leading zero for days 1 to 9; not a recording date. \\
\texttt{time\_of\_day\_code} & int & -- & Manual segment label: 0 for day and 1 for night. \\
\texttt{time\_of\_day\_name} & string & -- & Human-readable name corresponding to the time-of-day code. \\
\texttt{vehicle\_type\_code} & int & -- & Apparent or source-asserted recording-platform category code. \\
\texttt{vehicle\_type\_name} & string & -- & Human-readable name corresponding to the platform code. \\
\texttt{location} & object & -- & Nested locality, country, ISO3, continent, latitude, and longitude values. \\
\texttt{metadata} & object & -- & Limited structured per-upload provenance metadata copied from \texttt{mapping\_metadata.csv}; values may be null. Source titles, descriptions, and chapter text are excluded. \\
\texttt{mapping\_extra} & object & -- & Additional contextual fields copied from the source mapping row. \\
\texttt{mapping\_row\_number} & int & -- & Source row used to construct the segment record. \\
\bottomrule
\end{tabularx}
\end{table}


\subsection{Ethics statement}
CROWD is an index and derived pseudo-label release based on publicly viewable dashcam videos hosted on \textit{YouTube}. The project underwent institutional data protection review before large scale processing. The processing was conducted for scientific research using publicly available source material, in accordance with the applicable institutional data-protection framework and the research safeguards described in GDPR Article~89~\cite{gdpr2016}. Processing was limited to deriving contextual and object-level information; the footage was not combined with other datasets containing personal data, and no attempt was made to identify individuals. The underlying videos remain governed by the applicable YouTube terms and uploader-selected licence terms.

The public release does not redistribute source video, frames, audio, thumbnails, channel names, channel handles or custom channel URLs. It does include, in \texttt{mapping\_metadata.csv}, the title, description and chapter text of each upload as retrieved from the platform. This text was written by the uploaders, is not covered by the licence of the dataset, is not copied to \texttt{crowd\_index.jsonl}, and can contain names, links, contact details and other personal information. It retains platform-assigned video and channel identifiers, limited structured provenance metadata, segment timestamps, manual segment labels, contextual locality metadata, and derived YOLO/BoT-SORT detection and tracking pseudo-labels. The channel identifier is retained only to group uploads by source, assess source concentration, detect duplicate source associations, and audit provenance when uploads become unavailable. The derived outputs are limited to generic object classes, bounding boxes, confidence scores, frame indices, and segment-local track identifiers. They do not include licence-plate transcriptions, face or person identities, persistent identity labels, facial embeddings, biometric templates, walking-gait descriptors, or inferred sensitive personal attributes. Track identifiers are arbitrary within-segment object tracks and are not comparable across segments.

Public visibility does not remove privacy or data-protection obligations. Downstream users who retrieve or process the referenced videos are responsible for complying with applicable platform terms, copyright, privacy, data-protection, and research-ethics requirements. CROWD should not be used for identification, surveillance, biometric identification, face recognition, number-plate recognition, law-enforcement decisions, individual profiling, or persistent individual tracking. Users who store, redistribute, publish, or otherwise process source frames or clips should obtain any institutional review, lawful basis, permissions, and data-protection assessment required for their own use case.


\begin{table}[!htbp]
\centering
\setlength{\tabcolsep}{5pt}
\renewcommand{\arraystretch}{1.08}
\caption{Summary distribution of technical properties measured from the 82,427 downloaded source videos of the uploads in the dataset, analysed using \texttt{ffprobe}.
 Video share is calculated relative to all analysed source videos, while duration share is based on their source video durations rather than the retained CROWD segment duration. Only the derived technical property metadata and summary results are released; the source videos are not redistributed.}
\label{tab:video_quality_distribution}
\footnotesize
\begin{tabular}{llrrr}
\toprule
Property & Category & Videos & Video share (\%) & Duration share (\%) \\
\midrule
Exact resolution & 1280$\times$720 & 78,012 & 94.64 & 95.52 \\
Exact resolution & 960$\times$720 & 1,069 & 1.30 & 0.47 \\
Exact resolution & 640$\times$360 & 496 & 0.60 & 0.56 \\
Exact resolution & 1280$\times$676 & 317 & 0.38 & 0.50 \\
Exact resolution & 1270$\times$720 & 276 & 0.33 & 0.44 \\
Exact resolution & Remaining exact resolutions, each $<0.32\%$ & 2,257 & 2.74 & 2.51 \\
\midrule
Resolution bucket & 720p & 80,115 & 97.20 & 97.38 \\
Resolution bucket & 480p & 1,569 & 1.90 & 1.66 \\
Resolution bucket & 360p & 566 & 0.69 & 0.61 \\
Resolution bucket & 2160p & 121 & 0.15 & 0.29 \\
Resolution bucket & 240p & 47 & 0.06 & 0.04 \\
Resolution bucket & 1080p & 7 & 0.01 & 0.01 \\
Resolution bucket & 144p or lower & 1 & 0.00 & 0.00 \\
Resolution bucket & 1440p & 1 & 0.00 & 0.00 \\
\midrule
Frame rate & 30 fps & 35,923 & 43.58 & 41.26 \\
Frame rate & 29.97 fps & 24,687 & 29.95 & 31.22 \\
Frame rate & 25 fps & 5,242 & 6.36 & 7.50 \\
Frame rate & 60 fps & 4,877 & 5.92 & 6.89 \\
Frame rate & 59.94 fps & 3,932 & 4.77 & 4.91 \\
Frame rate & 24 fps & 2,186 & 2.65 & 1.64 \\
Frame rate & 23.98 fps & 1,826 & 2.22 & 2.81 \\
Frame rate & Non-standard frame rates, each $<0.47\%$ & 3,754 & 4.55 & 3.77 \\
\midrule
Aspect ratio & 16:9 & 78,634 & 95.40 & 96.38 \\
Aspect ratio & 4:3 & 1,225 & 1.49 & 0.59 \\
Aspect ratio & 320:169 & 319 & 0.39 & 0.50 \\
Aspect ratio & Remaining exact aspect ratios, each $<0.34\%$ & 2,249 & 2.73 & 2.53 \\
\midrule
Codec & h264 & 67,586 & 81.99 & 81.99 \\
Codec & vp9 & 10,888 & 13.21 & 13.54 \\
Codec & av1 & 3,953 & 4.80 & 4.48 \\
\midrule
Compression proxy & moderate bpppf & 65,185 & 79.08 & 75.90 \\
Compression proxy & low bpppf, strongly compressed & 16,005 & 19.42 & 22.43 \\
Compression proxy & high bpppf & 420 & 0.51 & 0.44 \\
Compression proxy & very low bpppf, highly compressed & 803 & 0.97 & 1.20 \\
Compression proxy & very high bpppf & 14 & 0.02 & 0.02 \\
\bottomrule
\end{tabular}
\end{table}
\section{Data records}
\label{sec:data}

The CROWD dataset is available through 4TU.ResearchData \cite{Alam2026CROWDdataset}. The deposited release contains four groups of data files: (i) the operational source tables \texttt{mapping.csv} and \texttt{mapping\_metadata.csv}; (ii) the derived segment-level index \texttt{crowd\_index.jsonl}; (iii) \texttt{crowd\_splits.csv}, with canonical splits and balancing weights for each segment; and (iv) per-segment CSV files containing frame-level YOLO11x detections linked by BoT-SORT. The source code used to create, process, and validate these files is provided separately as described in the Code Availability section.

The \texttt{mapping.csv} and \texttt{mapping\_metadata.csv} files are the operational tables used by the curation and processing code. The \texttt{mapping.csv} file has 10,538 rows, one per canonical locality, with an optional state or region for disambiguation. Its aligned upload lists contain 86,956 locality--upload associations representing 82,484 unique uploads; an upload can contribute to more than one locality record. Each row contains locality descriptors, geographic coordinates, contextual indicators, associated YouTube upload identifiers, segment boundaries, time-of-day labels, apparent recording-platform labels, and platform upload dates and platform-assigned channel identifiers where retrievable. Fields containing information for multiple uploads or segments are stored as aligned text-encoded lists. The schema is provided in \autoref{tab:mapping-schema}.

\begin{table}[!htbp]
\centering
\caption{Results of the automated release validation performed on the final frozen CROWD dataset.}
\label{tab:release-validation-results}
\small
\setlength{\tabcolsep}{5pt}
\renewcommand{\arraystretch}{1.05}
\begin{tabularx}{\textwidth}{X r l}
\toprule
Validation result & Value & Status \\
\midrule
Mapping rows checked & 10,538 & PASS \\
Unique source uploads referenced & 82,484 & PASS \\
Metadata rows checked & 82,484 & PASS \\
Uploads without complete platform metadata & 3,267 & WARN \\
JSONL records checked & 107,407 & PASS \\
JSONL invalid records & 0 & PASS \\
JSONL duplicate segment identifiers & 0 & PASS \\
JSONL missing segments & 0 & PASS \\
JSONL extra segments & 0 & PASS \\
JSONL records inconsistent with mapping & 0 & PASS \\
Achieved JSONL segment coverage (\%) & 100.00 & PASS \\
\midrule
Expected retained segments & 107,407 & PASS \\
Detection CSV files found & 107,407 & PASS \\
Detection CSV files matched to segments & 107,407 & PASS \\
Detection CSV files semantically checked & 107,407 & PASS \\
Detection rows semantically checked & All rows & PASS \\
Missing detection CSV files & 0 & PASS \\
Unmatched or unrecognised detection CSV files & 0 & PASS \\
Duplicate segment-to-file associations & 0 & PASS \\
Achieved detection segment coverage (\%) & 100.00 & PASS \\
\midrule
Invalid bounding-box rows & 0 & PASS \\
Exact duplicate detection rows & 0 & PASS \\
Non-monotonic frame-index rows & 0 & PASS \\
Rows with missing track identifiers & 0 & PASS \\
Inconsistent track-identifier rows & 0 & PASS \\
\midrule
Validator warning checks & 1 & WARN \\
Validator failures & 0 & PASS \\
\bottomrule
\end{tabularx}
\par\vspace{3pt}
\noindent\footnotesize\justifying Note: Both WARN entries come from the same validator warning, the coverage of the platform metadata. 3,267 of the 82,484 uploads do not have complete platform metadata: for 2,762 nothing was returned when the metadata was retrieved (consistent with uploads that are no longer publicly accessible), and 505 have a title, channel and view count but no upload date. This is reported as a warning and not as a failure because these uploads keep their segment boundaries, labels and detection files in \texttt{mapping.csv}.
\end{table}

The \texttt{mapping\_metadata.csv} file has 82,484 rows and is linked to \texttt{mapping.csv} through the YouTube video identifier. It contains complete limited structured provenance fields for 96.04\% of the 82,484 unique uploads, including the platform upload date, platform-assigned channel identifier, view count, number of retained segments, and date on which the record was last updated. Complete platform metadata is not available for 3,267 uploads (3.96\%): 2,762 for which nothing was returned when metadata retrieval was attempted (consistent with uploads that are no longer publicly accessible), and 505 with a title, channel and view count but no upload date (see the note to Table~\ref{tab:mapping-schema}). These uploads remain represented in \texttt{mapping.csv} and \texttt{crowd\_index.jsonl} because their source identifiers, segment boundaries, locality assignments, and manually curated labels are available; fields derived from \texttt{mapping\_metadata.csv} are empty for the affected records. The file also holds the title, description and chapter text of each upload as retrieved from the platform. This text is written by the uploaders, is not covered by the licence of the dataset, and is not copied to \texttt{crowd\_index.jsonl}.
The channel identifier is retained to group uploads originating from the same source and assess source concentration; it is not a channel name, handle, or custom URL. The platform upload date should not be interpreted as the recording date. The schema is provided in \autoref{tab:mapping-metadata-schema}.

The \texttt{crowd\_index.jsonl} file is a derived, segment-level representation of the structured information contained in the two operational CSV tables. It was added to provide a valid JSON representation of fields that are stored as text-encoded or nested lists in \texttt{mapping.csv}. Each non-empty line is an independent JSON object representing one retained segment. Nested values are represented using standard JSON objects and arrays, allowing users to read the index without applying Python-specific parsing to the CSV fields. Each object contains the source upload identifier and address, a unique segment identifier, segment start and end times, the manual time-of-day label, the apparent recording-platform label, the platform upload date where available, a nested locality object, limited structured provenance metadata, and the source mapping row number. Its schema is provided in \autoref{tab:jsonl-schema}.

The \texttt{crowd\_splits.csv} file has one row per segment of \texttt{crowd\_index.jsonl}, linked by \texttt{segment\_id}. It holds the video identifier, the channel group and the locality group used for splitting, the ISO3 code, the continent, the processed duration in seconds, the two splits (\texttt{split\_source} and \texttt{split\_locality}, each \texttt{train}, \texttt{val} or \texttt{test}) and the weights \texttt{weight\_locality} and \texttt{weight\_country}. Their use is described in the Usage Notes.

For each retained segment, the release provides a CSV file named \verb|{video_id}_{start_time}_{fps}.csv|. Each row contains an automatic object detection and its associated segment-local track identifier, confidence score, normalised bounding-box coordinates, object class, and frame index. The \verb|yolo-id| field contains the MS COCO class index produced by YOLO11x. Bounding boxes are represented by normalised centre coordinates, width, and height. The \verb|unique-id| field contains the BoT-SORT track identifier, which is unique only within the corresponding segment and cannot be compared across segments. These files contain automatic detection and tracking pseudo-labels and should not be interpreted as manually validated ground truth. Their schema is provided in \autoref{tab:yolo-schema}, and the tracking configuration is documented in \autoref{tab:tracking_hyperparams}.

\section{Technical validation}
\label{sec:technical_validation}

To ensure internal consistency and release integrity, we implemented an automated release \emph{validator} that operates directly on \texttt{mapping.csv}, \texttt{mapping\_metadata.csv}, \texttt{crowd\_index.jsonl}, and the per-segment detection CSV files. It checks structural correctness, cross-file consistency, segment coverage, duplicate associations, and row-level detection semantics. It does not constitute a manual assessment of the accuracy of the detector or tracker. The validator was run on the final frozen dataset with full row-level semantic checking enabled and completed without failures. It produced one coverage warning reporting the 3,267 uploads for which complete platform metadata was not available. The final results are summarised in \autoref{tab:release-validation-results}.

The final run checked all 10,538 locality records in \texttt{mapping.csv}, all 82,484 upload records in \linebreak \texttt{mapping\_metadata.csv}, and all 107,407 segment records in \texttt{crowd\_index.jsonl}. It also found and semantically checked all 107,407 per-segment detection CSV files. Each detection file was matched with its corresponding retained segment, with no missing, unmatched, or duplicate file associations. Both JSONL and detection-file segment coverage were 100\%. The 3,267 uploads without complete platform metadata do not reduce segment coverage because the affected uploads retain their segment definitions and manual annotations in \texttt{mapping.csv}.

\paragraph{Screen and manual review of candidate discontinuities.}
A retained segment is meant to be one continuous, unedited stretch of driving, so after the first, manual screening the segments were checked a second time for discontinuities that may have been missed. The idea of this second step is to combine an exhaustive automatic scan with a human decision. Watching all of the footage again is not feasible, but a detector of abrupt visual change can point to the few seconds in which a discontinuity is likely. A detector alone is not reliable enough to cut the data, however: whether a sudden change is an edit, or the entrance of a tunnel, a flash of sunlight, a passing lorry or a camera adjustment, is a judgement that needs context. We therefore let the detector propose candidates and a reviewer decide, and only decisions made by a person were applied. The numbers below support this choice: of the candidates that were reviewed, most were not cuts, so splitting at every candidate would have broken many continuous segments.

The screen ran FFmpeg scene-change detection (pixel-level scene score; threshold 0.45) on the original video file of each segment, over the retained interval only. Flagged times closer than 2.0 s were merged into one candidate, and each candidate was recorded as an absolute time in the original upload, together with the segment start and the frame rate. Segments that had been processed before were not processed again. The screen flagged 52,838 candidate cuts in total. The threshold of 0.45 favours abrupt cuts. Gradual transitions such as cross-fades, and cuts between visually similar scenes, can be missed, and the screen does not detect changes of playback speed or advertisements as such.

The candidates were reviewed by a member of the author team in a web-based tool that plays the source upload from two seconds before to two seconds after the candidate. The reviewer recorded whether the candidate was a genuine cut (\emph{yes}) or not a cut (\emph{no}), and could add a note; a candidate in an upload that could no longer be played in the embedded player is counted as not a cut. In total, 7,347 candidates (13.9\% of those flagged) were reviewed, in the order in which they were presented and not as a random sample, so the shares below describe the reviewed candidates only. Of these, 2,481 (33.8\%) were genuine cuts and 4,866 (66.2\%) were not. The 2,481 confirmed cuts lie in 1,363 segments, which is 40.7\% of the 3,346 segments with at least one reviewed candidate. The review covered the segments of uploads added by all curators, including those added by the other authors. Besides cuts, the reviewer looked for two further problems that the screen does not find: footage that is not at normal ($1\times$) playback speed, for example fast-forwarded passages, and advertisements inserted in the middle of a video. These were looked for while watching the candidates, so they were checked in the four seconds around each flagged time and not along the whole segment, and segments whose candidates were not reviewed may still contain such footage.

A confirmed cut, or an advertisement or a passage that is not at normal speed, was removed from the retained segments by splitting the segment around it, or by removing the segment where it affected the whole segment. For a confirmed cut at time $t$, the segment is split into a part that ends at $\lfloor t\rfloor$ and a part that starts at $\lceil t\rceil$, so that one to two seconds around the cut are dropped, and parts shorter than one second are discarded. The per-segment detection file was split at the same frames, with frame indices restarted at 1 for each new segment. The true frame rate read from the video file was used, not the rounded rate in the file name, so that the split files stay aligned with the new segment boundaries.

\paragraph{Integrity of the source videos.}
The downloaded video files were inventoried with \texttt{ffprobe} to find files without a readable video stream (broken or partial downloads), duplicate files, files that were more than five seconds shorter than the end of their last segment (cut off while downloading, or with an end time beyond the end of the video), and files that are no longer referenced by any mapping file. Affected videos were downloaded again, or the affected segments were trimmed to the available length, together with their detection files.

\paragraph{Audit of locality coordinates and populations.}
In September 2026, the coordinates and populations of the localities were checked against GeoNames and, where they disagreed, against Wikidata and OpenStreetMap. In total, 484 populations were corrected, mostly values copied from a same-named place or from another row, district or county figures given for single towns, and impossible values. Twelve localities that had been geocoded to a same-named place in another region were moved, wrong states and invalid upload dates were corrected where they could be determined, and zeros that meant ``no data'' were cleared. At the time of the audit, 1,283 localities could not be matched to GeoNames, so their populations remain unchecked.

\paragraph{Quantitative audit of labels and segment boundaries.}
We audited a stratified random sample of 225 segments, drawn at random by continent and time-of-day label, with at most one segment per upload: 25 in each of the strata of Africa, Asia and Europe (day and night) and of North America (day), 15 in North America (night), 12 in Oceania (day), 10 in Oceania (night), 8 in South America (day) and 5 in South America (night). An author who had not added the upload re-watched each sampled segment and recorded whether the start and end times were correct (and the corrected times if not), the time of day and the platform type, without seeing the labels of the dataset. Oceania and South America have fewer segments in the sample, so their intervals are wide. The boundaries of 0 of the 225 segments (0.0\%; 95\% Wilson interval 0.0 to 1.7\%) had to be corrected, the time-of-day label differed for 0 (0.0\%; 0.0 to 1.7\%; Cohen's $\kappa=1.00$), and the platform label differed for 1 (0.4\%; 0.1 to 2.5\%), a segment in South America that the dataset labels as a car and the reviewer as an emergency vehicle.

\paragraph{Manual evaluation of detections and tracks.}
To establish the range of quality of the pseudo-labels, we checked them by hand on a stratified random sample of footage. The sample has 480 frames (40 for each continent and time-of-day stratum, at most one per upload, drawn with a probability proportional to the footage), and, for tracking, 118 tracks from the 60 segments of a second sample (five segments per stratum, two tracks per segment). One author reviewed each frame with the boxes of the detector at the confidence threshold of 0.7 that the analyses of the dataset use, in a browser-based tool that is part of the repository (\texttt{release/annotate.py}). The reviewer marked every box as correct, as a false positive (no such object), or as the wrong class (and gave the true class), and drew a box around every person, bicycle, car, motorcycle, bus, truck, traffic light or stop sign that had no box (a missed object). A correct box is a true positive; a false positive, and a box of the wrong class, is a false positive of the predicted class; a missed object, and the true class of a wrongly classified box, is a false negative. \autoref{tab:detection_eval} reports precision, recall and F1 per class, with 95\% bootstrap intervals over frames in the repository. All 480 frames were reviewed; they have 1,296 boxes at the threshold, all of which the reviewer marked as correct, and 423 missed objects, so that precision is 1.00 in every class and the differences between classes are in the recall. The recall is 0.75 over the eight classes (95\% interval 0.72 to 0.79), and lowest for traffic lights (0.31), motorcycles (0.58) and trucks (0.60). The evaluation covers eight classes: the six road users that the analyses of the dataset use (person, bicycle, car, motorcycle, bus and truck), and traffic lights and stop signs; the other 72 classes of the detector were not evaluated. We do not report an average precision, because only the boxes above the threshold were shown. Showing the boxes of the detector is faster than drawing every box, but it can make a reviewer accept them too readily, so precision may be somewhat optimistic.
Results per time of day and continent are given in the repository. For tracking, we asked one question per track: is the box on one and the same object all the time? The reviewer saw a short video of the track, zoomed on the object, with only the box of that track drawn (up to three pieces of 5 s, from the start, the middle and the end of the track), and answered that the track follows one object, that the box jumps to another object (an identity switch), that the box is not on a road user (a false track), or that this cannot be told. The tracks were drawn at random from the road users that the tracker followed for at least 2 s at a confidence of 0.7, two for each of the 60 segments. Of the 118 tracks, 115 followed one object, 1 had an identity switch (0.9\% of the 116 tracks on a road user; 95\% Wilson interval 0.2 to 4.7\%), and 2 were not on a road user (1.7\%; 0.5 to 6.0\%). This is an error taxonomy per track and not a standard tracking score such as IDF1 or MOTA, because identities were not annotated, and it does not measure fragmentation, that is, one object split over several tracks, nor the objects that were missed; the missed objects are counted in the recall of the table. Video quality was not a stratum of the sample; the frames can be joined to the \texttt{ffprobe} descriptors of the videos to report quality-specific results.

\begin{table}[!htbp]
\centering
\caption{Quality of the detections on the audit frames, from a manual review of the boxes at a confidence of 0.7 (precision, recall and F1).}
\label{tab:detection_eval}
\small
\begin{tabular}{lrrrr}
\toprule
Class & Objects & Precision & Recall & F1 \\
\midrule
Person & 223 & 1.00 & 0.77 & 0.87 \\
Bicycle & 6 & 1.00 & 0.83 & 0.91 \\
Car & 1,115 & 1.00 & 0.83 & 0.91 \\
Motorcycle & 67 & 1.00 & 0.58 & 0.74 \\
Bus & 43 & 1.00 & 0.81 & 0.90 \\
Truck & 128 & 1.00 & 0.60 & 0.75 \\
Traffic light & 132 & 1.00 & 0.31 & 0.47 \\
Stop sign & 5 & 1.00 & 0.80 & 0.89 \\
All eight classes & 1,719 & 1.00 & 0.75 & 0.86 \\
\bottomrule
\end{tabular}
\end{table}

\paragraph{Downloaded video format and compression descriptors.}
Because video technical properties can affect perception performance, we characterised the 82,427 downloaded source video files of the dataset, one for each upload except 57 for which no file was available. For each file, we extracted the first video stream metadata using \texttt{ffprobe}, including frame width, frame height, average frame rate, codec, pixel format, duration, and file size. All 82,427 files were probed successfully, with no failed probes. \autoref{tab:video_quality_distribution} reports the directly observed distributions by resolution, frame rate, aspect ratio, codec, and compression proxy. Video share is calculated relative to all 82,427 analysed files, while duration share is calculated from their source video durations.

Because CROWD is sourced from YouTube, the original camera recordings prior to YouTube encoding and compression were unavailable. The downloaded files were already compressed platform versions and therefore could not be used as references for calculating reference based perceptual quality metrics. We instead report bits per pixel per frame (bpppf) as an approximate compression proxy, computed as effective bitrate divided by frame width $\times$ frame height $\times$ frame rate. Effective bitrate was calculated as file size $\times 8$ divided by duration, rather than relying on stream level bitrate fields that may be missing or implausible. We grouped this proxy into five broad categories: very low bpppf ($<0.03$), low bpppf ($0.03$ to $<0.06$), moderate bpppf ($0.06$ to $<0.12$), high bpppf ($0.12$ to $<0.25$), and very high bpppf ($\ge 0.25$). This measure should be interpreted as a technical descriptor of compression level rather than as perceptual quality ground truth. Across the 82,427 analysed files, the median effective bitrate was 2.00 Mbps (interquartile range 1.78 to 2.14 Mbps), and the median bpppf was $7.19*10^{-2}$ (interquartile range $6.28*10^{-2}$ to $7.74*10^{-2}$).

\begin{table}[!htbp]
\centering
\caption{Schema of per segment YOLO detection and BoT-SORT tracking pseudo-label CSV files named \texttt{\{video\_id\}\_\{start\_time\}\_\{fps\}.csv}. Column names follow the CSV header.}
\begin{tabular}{llll}
\toprule
Field & Type & Units & Notes \\
\midrule
\texttt{yolo-id} & int & -- & COCO class index, from 0 to 79. \\
\texttt{x-center} & float & -- & Box centre x, normalised by image width, in $[0,1]$. \\
\texttt{y-center} & float & -- & Box centre y, normalised by image height, in $[0,1]$. \\
\texttt{width} & float & -- & Box width, normalised by image width, in $[0,1]$. \\
\texttt{height} & float & -- & Box height, normalised by image height, in $[0,1]$. \\
\texttt{unique-id} & int & -- & BoT SORT track identifier, unique within the segment, $\ge 0$. \\
\texttt{confidence} & float & -- & Detector confidence score in $[0,1]$. \\
\texttt{frame-count} & int & frames & Frame index within the segment, first frame is 1, $\ge 1$. \\
\bottomrule
\end{tabular}

\label{tab:yolo-schema}
\end{table}


\paragraph{Internal consistency of the mapping table.}
The validator parses the mapping table and verifies that the list encoded fields are structurally aligned. For each mapping row, it checks that the outer list lengths describing videos and their associated attributes, for example time of day labels, platform labels, and segment boundary lists, are consistent. It also checks that categorical fields use the permitted values. Within each video entry, it validates segment boundaries by requiring finite numeric timestamps with non negative values and strictly ordered intervals, with start $<$ end. These checks detect mis defined rows, misaligned list encodings, invalid categorical values, and invalid segment definitions; they confirm structural completeness of the released labels but do not turn the labels into fine grained perception ground truth.

To eliminate duplicate locality records not captured by name matching, we also calculated pairwise distances from the recorded latitude and longitude and manually reviewed locality pairs within 2 km. When two records were confirmed to represent the same locality, their mapping rows were merged, with one name retained in \texttt{locality} and the alternative name added to \texttt{locality\_aka}. Nearby records representing distinct localities were retained separately. In addition, towns of fewer than 250,000 inhabitants that lie within 30 km of a city of more than one million inhabitants in the same state or region were shown side by side on maps and merged into that city when the reviewer judged them to belong to its metropolitan area.

\paragraph{Recomputation of paper aligned dataset summaries.}
Using the validated mapping content, the validator recomputes the primary descriptive statistics reported in the manuscript: the number of unique uploads, the number of upload records, the number of segment records, and the total annotated duration obtained by summing the adjusted processing interval, $t_{\mathrm{end,processed}}-t_{\mathrm{start}}$, across segments. It also reproduces continent level aggregations of segment records and duration, continent wise day and night label entry distributions, and global as well as continent stratified upload composition, day only, night only, and both. Where reference totals are available, the recomputed values are compared with those references and any discrepancies are flagged.

\paragraph{Validation of the JSON Lines index.}
The validator reads \texttt{crowd\_index.jsonl} line by line and verifies that every non-empty line is valid JSON, that every record is an object with a unique non-empty \texttt{segment\_id}, and that the required top-level and nested location fields are present. Each JSONL segment is compared with the corresponding segment derived from \texttt{mapping.csv}, including the upload identifier, start and end times, source mapping row, time-of-day code, and apparent platform code. The final run parsed all 107,407 JSONL records successfully. It found no invalid JSON lines, duplicate segment identifiers, missing required fields, missing or extra segments, or values inconsistent with the corresponding mapping records. The achieved JSONL segment coverage was 100\%.

\paragraph{Cross file consistency between mapping and per video metadata.}
The validator performs bidirectional consistency checks between the mapping table and the per video metadata table. It reports mapping-table uploads without a corresponding metadata record and metadata records whose upload identifier is absent from the mapping, excluding placeholder identifiers. The final run found a metadata row for all 82,484 unique mapping-table uploads and 3,267 metadata rows with missing platform fields; the result is reported as a coverage warning rather than a failure of the retained segment annotations. The stored per-video \texttt{segments} value in \texttt{mapping\_metadata.csv} is a non-authoritative summary and is not validated; the retained segment inventory is derived from \texttt{mapping.csv} and \texttt{crowd\_index.jsonl}.

\paragraph{Integrity and coverage of detection outputs.}
The validator scans the folder of detector produced CSV files and enforces a naming convention that encodes the source video identifier, segment start time, and frame rate. For each detection file, it verifies that the referenced video identifier exists in the mapping table and that the encoded segment start time corresponds, within a fixed tolerance, to a known segment start time for that video derived from the mapping table. The final run found 107,407 detection CSV files and matched every file to one of the 107,407 expected retained segments. It found no missing files, unmatched or unrecognised files, or duplicate segment-to-file associations, yielding 100\% detection-file segment coverage.

\paragraph{Semantic validation of detection and tracking rows.}
For the full release run, the validator opened all 107,407 uniquely matched detection CSV files and checked that all required columns were present and readable. For every detection row, it verified finite and permitted class identifiers, confidence values in $[0,1]$, positive normalised box dimensions, box extents within the image bounds, and valid positive frame indices. It also checked for exact duplicate rows, decreases in frame index in file order, missing or invalid track identifiers, changes in object class within a track, and reuse of the same track identifier more than once in the same frame. The validator found zero invalid bounding-box rows, zero exact duplicate rows, zero non-monotonic frame-index rows, zero rows with missing track identifiers, and zero inconsistent track-identifier rows.


\begin{table}[!htbp]
\centering
\setlength{\tabcolsep}{4pt}
\renewcommand{\arraystretch}{1.0}
\caption{BoT-SORT tracking hyperparameters used to generate segment level track identifiers in the released per segment CSV files.}
\label{tab:tracking_hyperparams}
\vspace{-4pt}
\begin{tabularx}{\textwidth}{l l X}
\toprule
Hyperparameter & Value & Description \\
\midrule
\texttt{tracker\_type} & \texttt{botsort} & Tracker implementation selected by the custom YAML configuration. \\
\texttt{track\_high\_thresh} & 0.70 & Confidence threshold for the first association stage during tracking. \\
\texttt{track\_low\_thresh} & 0.30 & Confidence threshold for the second association stage during tracking. \\
\texttt{new\_track\_thresh} & 0.70 & Minimum confidence required to initialise a new track when no matches are found. \\
\texttt{track\_buffer} & $2 \times \mathrm{FPS}$ frames & Frame-rate-dependent runtime value corresponding to two seconds at the rounded frame rate of the segment. The YAML value of 60 at the 30 fps reference rate is overwritten before segment processing. \\
\texttt{match\_thresh} & 0.60 & Similarity threshold used to match existing tracks with detections. \\
\texttt{fuse\_score} & \texttt{true} & Combines detection confidence with IoU distance for matching. \\
\texttt{gmc\_method} & \texttt{sparseOptFlow} & Global motion compensation method used for moving camera footage. \\
\texttt{proximity\_thresh} & 0.50 & IoU threshold for spatial proximity during association. \\
\texttt{appearance\_thresh} & 0.25 & Threshold for appearance based matching. \\
\texttt{with\_reid} & \texttt{true} & Enables appearance based re identification. \\
\texttt{model} & \texttt{auto} & No separate ReID weights were specified; resolution of \texttt{auto} followed Ultralytics 8.4.46. \\
\bottomrule
\end{tabularx}
\end{table}

\section{Usage Notes}

The dataset distributes the structured \texttt{crowd\_index.jsonl} index, the source CSV tables, platform-assigned video and channel identifiers, segment start and end times, limited structured provenance metadata, and derived detection and tracking pseudo-labels, but it does not redistribute the underlying video, frames, audio, thumbnails, or channel names or handles. The title, description and chapter text of each upload are included in \texttt{mapping\_metadata.csv} as retrieved from the platform; they are third-party text and are not covered by the licence of the dataset. Users can access referenced uploads through YouTube using the provided source addresses or video identifiers. For analyses that require local video processing, we provide the code used in this work to retrieve and preprocess the referenced uploads in the Code Availability section; use of this code is subject to platform terms and video availability. Because the referenced videos are hosted by YouTube, some uploads and metadata may become unavailable or change over time due to removal, restriction, regional limitations, or account changes.


\paragraph{Software, device, and deterministic execution.}
The supporting package versions were torchvision 0.23.0, NumPy 1.26.4, SciPy 1.11.4, \texttt{lap} 0.5.12, \texttt{filterpy} 1.4.5, and PyYAML 6.0.3. Video retrieval and inspection used \texttt{pytubefix} 10.10.1, \texttt{yt-dlp} 2026.7.4, and FFmpeg and \texttt{ffprobe} 4.4.2. PyTorch was built with CUDA 12.8 support, and the inspected environment reported cuDNN 9.10.2. The complete installed package manifest is provided with the reproducibility outputs.

The processing code selects CUDA when available and otherwise falls back to CPU. The generated environment report for the inspected machine records an NVIDIA GeForce RTX 4080 SUPER GPU with 16 GB of memory and NVIDIA driver 560.35.05. This machine record is not presented as the hardware used for every historical processing job because the GPU model was not contemporaneously logged for all earlier runs. The Ultralytics configuration used seed 0 and enabled its deterministic option. However, PyTorch deterministic algorithms and deterministic cuDNN execution were not enabled, and determinism related environment variables, including \texttt{CUBLAS\_WORKSPACE\_CONFIG} and \texttt{PYTHONHASHSEED}, were unset. The processing pipeline is therefore not claimed to be bitwise deterministic across hardware or software environments.

\paragraph{Privacy and data protection.}
CROWD should be treated as a privacy-sensitive research index because the provided source identifiers and segment boundaries may allow users to access videos containing identifiable people, vehicles, residences, and locations. The released files do not contain video frames, audio, thumbnails, face identities, number-plate transcriptions, persistent person identities, facial embeddings, biometric templates, or walking-gait descriptors. Platform-assigned channel identifiers are retained only for grouping uploads by source, assessing source concentration, detecting duplicate source associations, and auditing provenance; they should not be used to identify or profile uploaders. The metadata table does contain the titles, descriptions and chapter text written by uploaders, which can include names, links or contact details; this text should not be used to identify or profile uploaders and should be handled in accordance with applicable data protection requirements. Downstream users who retrieve the referenced videos create their own processing context and are responsible for complying with platform terms, copyright, privacy, data protection, and research ethics requirements. We recommend data minimisation, restricted access to any locally retrieved video, avoidance of individual identification, and task-specific institutional review where source footage or derived imagery is processed.

Users should be aware of several limitations relevant to reuse. The object detections and tracks are automatic pseudo-labels generated from a detector trained on the MS COCO label set and therefore cover only the corresponding classes (e.g. persons, bicycles, motorcycles, cars, buses, trucks, traffic lights, and stop signs). They are not manually verified annotations (their accuracy on a stratified sample is reported in the Technical validation) and should not be treated as ground truth for object detection or tracking accuracy. These pseudo-labels do not include lane markings, road geometry, or a comprehensive traffic sign taxonomy beyond the COCO categories. Detection and tracking quality may degrade under occlusion, dense traffic, motion blur, low illumination, adverse weather, unusual camera viewpoints, and domain shifts across countries or localities. Track identifiers are not propagated across segment boundaries because tracking is reinitialised per segment. Users requiring ground truth labels should validate or manually annotate the relevant subset for their task.

The coverage of time-of-day is uneven: most retained content is driving during the day (Table~\ref{tab:continent-daynight-uploads}). Analyses or models intended for nighttime conditions should therefore account for this imbalance (for example, by stratified sampling, weighting, or separate evaluation by time-of-day label). CROWD also separates the platform upload date from the recording date: only upload dates are provided, and no independently verified recording dates are released. Users should therefore avoid interpreting \autoref{fig:upload_month} as a temporal distribution of driving exposure or using upload dates directly for policy, seasonal, or Covid-19 analyses unless they add independently supported recording date evidence with explicit provenance and uncertainty.

\paragraph{Sampling design and cross-country comparisons.}
Because CROWD is a convenience sample (see the Methods), the amount of footage of a country or locality is not a measure of its driving exposure, and differences between places can reflect who uploads footage, which channels dominate, and which vehicles and times of day are filmed, instead of differences in behaviour. For comparisons between countries we recommend: (i) comparing within strata of time of day and platform type, because they differ strongly between countries; (ii) setting a minimum amount of footage, so that an estimate is not based on a handful of uploads; (iii) weighting by the balancing weights described below, or resampling, so that high-volume places do not decide the result; and (iv) quantifying uncertainty by resampling uploads or, better, channels, not segments, because segments of one source are not independent. \autoref{tab:min_duration} shows how the choice of a minimum duration changes the sample. A minimum of 5 h per locality leaves 520 localities (4.9\% of all localities) in 121 countries that hold 84.7\% of the footage, and a minimum of 10 h per country leaves 109 countries (99.0\% of the footage); we suggest these as starting points, not as validated thresholds.

\begin{table}[!htbp]
\centering
\caption{Effect of a minimum amount of processed footage on the sample. Shares are of all localities (or countries and territories) and of all footage.}
\label{tab:min_duration}
\small
\begin{tabular}{rrrrr}
\toprule
\multicolumn{5}{@{}l}{\textit{(a) Minimum footage per locality}} \\
Minimum (h) & Localities & Share of localities (\%) & Countries covered & Share of footage (\%) \\
\midrule
1 & 2,043 & 19.4 & 181 & 93.2 \\
5 & 520 & 4.9 & 121 & 84.7 \\
10 & 297 & 2.8 & 92 & 80.5 \\
50 & 107 & 1.0 & 52 & 68.8 \\
100 & 72 & 0.7 & 38 & 61.9 \\
\midrule
\multicolumn{5}{@{}l}{\textit{(b) Minimum footage per country or territory}} \\
Minimum (h) & Countries & Share of countries (\%) & -- & Share of footage (\%) \\
\midrule
10 & 109 & 45.8 & -- & 99.0 \\
100 & 53 & 22.3 & -- & 93.0 \\
1,000 & 8 & 3.4 & -- & 56.1 \\
\bottomrule
\end{tabular}
\end{table}

\paragraph{Balanced evaluation protocols.}
Footage is highly concentrated: the top 1\% of localities hold 68.7\% of the hours, and 52\% of the localities consist of a single segment. A model trained or evaluated on randomly drawn segments is therefore mostly trained or evaluated on a few large cities. We recommend one of three protocols. In a \emph{locality-capped} protocol, at most a fixed amount of footage per locality is used (\autoref{tab:cap}); a cap of 10 h, for example, keeps 27.7\% of the footage and reduces the share of the top 1\% of localities from 68.7\% to 10.6\%. In a \emph{locality-balanced} or \emph{country-balanced} protocol, segments are drawn with probabilities proportional to the columns \texttt{weight\_locality} or \texttt{weight\_country} of \texttt{crowd\_splits.csv}, which sum to one within each locality or country, so that every locality or country contributes equally in expectation; the country-balanced draw should be restricted to countries above a minimum duration (\autoref{tab:min_duration}). Results should be reported both on the unweighted data and under at least one balanced protocol.

\begin{table}[!htbp]
\centering
\caption{Locality-capped sampling: the footage that remains when each locality contributes at most the stated number of hours.}
\label{tab:cap}
\small
\resizebox{\linewidth}{!}{%
\begin{tabular}{rrrrr}
\toprule
Cap per locality (h) & Remaining footage (h) & Share of footage (\%) & Localities at the cap & Share of the top 1\% of localities (\%) \\
\midrule
5 & 8,147 & 22.5 & 520 & 6.5 \\
10 & 10,029 & 27.7 & 297 & 10.6 \\
25 & 13,272 & 36.7 & 166 & 20.0 \\
50 & 16,645 & 46.0 & 107 & 31.8 \\
\bottomrule
\end{tabular}%
}
\end{table}

\paragraph{Canonical splits and source leakage.}
Several segments can come from one upload, an upload can be associated with more than one locality, and the uploads of one channel share cameras, routes, compression and uploader-specific content. A random split of segments therefore leaks source information between the training and test data. We provide canonical splits in \texttt{crowd\_splits.csv}. In \texttt{split\_source}, the groups are source channels (an upload without a known channel is its own group), so no channel, and therefore no upload, appears in more than one split. In \texttt{split\_locality}, the groups are localities, so a locality is never shared, which tests generalisation to unseen places. In both, a group is assigned to the training (70\%), validation (15\%) or test (15\%) split by a hash of its name, so the assignment is reproducible. We verified that none of the 5,134 channel groups and none of the uploads is in more than one split of \texttt{split\_source}, and that none of the 10,538 localities is in more than one split of \texttt{split\_locality}. Because a few channels and localities are very large, the shares of footage deviate from the nominal ones (\autoref{tab:splits}). We recommend reporting results on \texttt{split\_source} as the default, on \texttt{split\_locality} for geographic generalisation, and grouping by \texttt{channel\_group} in any cross-validation.

\begin{table}[!htbp]
\centering
\caption{Canonical splits. Uploads can occur in more than one split of \texttt{split\_locality}, because an upload can be associated with several localities.}
\label{tab:splits}
\footnotesize
\setlength{\tabcolsep}{4pt}
\begin{tabular}{lrrrrrr}
\toprule
Split & Segments & Footage (h) & Uploads & Groups & Localities & Countries \\
\midrule
\multicolumn{7}{@{}l}{\textit{split\_source (groups are channels)}} \\
Train & 69,172 (64.4\%) & 22,201 (61.3\%) & 52,763 & 3,618 & 7,630 & 231 \\
Val & 16,516 (15.4\%) & 5,400 (14.9\%) & 13,205 & 733 & 2,789 & 155 \\
Test & 21,719 (20.2\%) & 8,603 (23.8\%) & 16,516 & 783 & 2,380 & 157 \\
\midrule
\multicolumn{7}{@{}l}{\textit{split\_locality (groups are localities)}} \\
Train & 74,900 (69.7\%) & 25,583 (70.7\%) & 57,940 & 7,328 & 7,328 & 226 \\
Val & 18,944 (17.6\%) & 6,371 (17.6\%) & 15,691 & 1,627 & 1,627 & 161 \\
Test & 13,563 (12.6\%) & 4,251 (11.7\%) & 10,510 & 1,583 & 1,583 & 164 \\
\bottomrule
\end{tabular}
\end{table}

\paragraph{Availability of the source videos and versioning.}
The videos remain on YouTube, so the visual corpus that a user can rebuild can shrink or change after a release, even when the derived index is unchanged. We therefore treat the availability of the sources as a property of the release. We recommend the following procedure for disappeared sources. (i) The segment records, labels and detection files of a source are never deleted when the source disappears, so that results remain reproducible from the index. (ii) Users cite the release (version and date) and report the share of segments whose video they could retrieve. (iii) Results are reported on all segments of the index when the analysis uses only the derived files, and on the retrievable subset when it uses pixels. (iv) A retrieved video is verified against the dataset: its duration and frame rate should match the processed duration and the frame rate in the name of the detection file, and the last frame index of the detection file should be close to the processed duration times the frame rate. (v) Re-encoded copies, with a different resolution or codec, can shift frames slightly, so a user who needs exact alignment should re-run the detector on the retrieved copy and compare. Rows with missing platform fields mark the sources that could not be retrieved, which is only a proxy for sources that are no longer publicly accessible.
\paragraph{File storage.}
Files larger than 99 MB are stored in parts (\texttt{NAME\_part1.EXT}, \texttt{NAME\_part2.EXT}, \ldots) because of the file size limit of the hosting service; scripts in the repository join them again.

\paragraph{Time-of-day and platform labels.}
The \texttt{time\_of\_day} field provides one binary contextual label for every retained segment. Its operational definition, intended scope, and limitations are described earlier in the Methods section. The field supports broad filtering, descriptive stratification, and dataset characterisation. Users requiring finer illumination categories or uncertainty-aware labels should independently relabel or audit the relevant subset for their task.


\begin{table}[!htbp]
\centering
\setlength{\tabcolsep}{4pt}
\renewcommand{\arraystretch}{1.0}
\caption{Pinned software environment and effective inference settings used to generate the released per-segment detection and tracking pseudo-label files.}
\label{tab:detection_environment}
\vspace{-4pt}
\begin{tabularx}{\textwidth}{l l X}
\toprule
Item & Value & Meaning \\
\midrule
Python & 3.12.13 &
Python runtime used to execute the detection and tracking pipeline. \\

Ultralytics & 8.4.46 &
Software package used to load the YOLO11x model and run detection and tracking inference. \\

PyTorch & 2.8.0+cu128 &
Deep learning framework used to execute the YOLO11x neural network. \\

OpenCV & 4.11.0.86 &
Library used for video decoding, frame handling, and related image operations. \\

Image size & 640 &
Nominal inference input size passed to Ultralytics. Video frames were internally resized and padded for model inference while preserving their aspect ratio. \\

NMS IoU threshold & 0.70 &
Overlapping detections of the same class were suppressed when their intersection over union exceeded this threshold during non maximum suppression. \\

Detector confidence threshold & 0.00 &
Minimum confidence threshold passed to Ultralytics during YOLO11x inference. Subsequent detection association used the BoT-SORT thresholds reported separately. \\

Maximum detections & 300 &
Maximum number of object detections retained for any processed frame after non maximum suppression. \\

Class agnostic NMS & false &
Non-maximum suppression was performed separately for each object class, so boxes assigned to different classes did not suppress one another. \\

Class filter & none &
No object classes were excluded during inference; detections from all 80 MS COCO classes were eligible for retention. \\

Rectangular inference & false &
Standard square inference batches were used rather than minimum rectangle batching based on the aspect ratios of the input frames. \\

Half precision & false &
Inference used full 32 bit floating point calculations rather than 16 bit floating point calculations. \\

DNN inference & false &
The OpenCV DNN inference backend was not used; model inference was performed through PyTorch. \\

Augmentation & false &
Each frame was processed once without test time augmentation such as scaling or flipping. \\

Video stride & 1 &
Every frame in each retained segment was processed, with no systematic frame skipping. \\

\bottomrule
\end{tabularx}
\end{table}


The \texttt{vehicle\_type} field should likewise be interpreted as an apparent recording platform or a category of a confirmed platform source, not as an independently verified platform identity. The appearance of the platform can be uncertain in forward-facing video sourced from the web because the height of the camera, the mounting geometry, the field of view, the reflections, the shadows, and the source descriptions may be incomplete or ambiguous. Automated driving status is not independently verified in CROWD; automated variants were assigned only when the title, description, or chapter information provided by the source indicated automated driving or autonomous shuttle operation and should be treated as apparent or asserted by the source unless a downstream user independently verified them.

Geographic coverage is also uneven. Africa is the second largest continent by land area and the second most populous continent after Asia, yet it contributes the smallest retained duration in our collection (Table~\ref{tab:continent-segments}). The concentration at the country and locality level is also substantial (Table~\ref{tab:country-locality-concentration}): the ten largest contributing countries account for 60.47\% of retained hours, while the top 1\%, 5\% and 10\% of localities account for 68.66\%, 84.78\% and 89.35\% of retained hours, respectively. Studies that aim to make claims across regions should therefore consider continent, country, and locality specific sampling strategies, minimum duration thresholds, weighting, and stratified reporting.

\section{Data Availability}
\label{sec:data_avail}
The CROWD dataset is available via 4TU.ResearchData (\url{https://doi.org/10.4121/06e9bb9a-a064-412b-b0f3-9ac5dd62ea16}). The deposit includes \texttt{crowd\_index.jsonl}, \texttt{mapping.csv}, \texttt{mapping\_metadata.csv}, \texttt{crowd\_splits.csv}, and the per-segment detection and tracking CSV files. The deposited index files, manual labels, contextual and limited structured provenance metadata, derived detection and tracking pseudo-labels, and validation outputs are released under a CC BY 4.0 data licence. The titles, descriptions and chapter text in \texttt{mapping\_metadata.csv} are third-party text written by the uploaders. They and the underlying YouTube videos (which are not redistributed) are not covered by this licence, and remain governed by platform terms, copyright, privacy, and data protection requirements, and the uploader-selected licence terms. The source code and configuration files are provided separately under the MIT License, as described in the Code Availability section.


\section{Code Availability}
\label{sec:code_avail}
The source code to download videos, curate segments, generate \texttt{crowd\_index.jsonl}, produce the per-segment pseudo-label CSVs, and run the release validator is available at \url{https://github.com/crowd-dataset/crowd} and is released under the MIT License. The repository contains the JSONL conversion script, \texttt{validation\_release.py}, the release configuration files (e.g. \texttt{config}, \texttt{bbox\_custom\_tracker.yaml}), and an environment specification (\texttt{pyproject.toml}) targeting Python 3.12.13. The tracking of segments that are processed in parallel threads also uses a per-thread track identifier counter (\texttt{track\_ids.py}, \url{https://github.com/crowd-dataset/crowd/blob/main/track_ids.py}), so that every segment is numbered independently. The reproducibility outputs include \texttt{reproducibility\_report.json}, the complete installed package manifest \texttt{pip\_freeze.txt}, the model and tracker checksums, the effective inference settings, the tracker configuration, and the recorded CUDA, cuDNN, GPU, and deterministic settings. The validator also regenerates the machine-readable validation report, detailed issue table, manuscript-ready validation table, and checksum manifest. The repository also contains the script that creates the splits and weights (\texttt{release/make\_splits.py}), the scripts that draw and score the manual audits (\texttt{release/sample\_audit.py}, \texttt{release/audit\_stats.py} and \texttt{release/evaluate\_detections.py}), and the tools used to find and add videos (the form-based ingestion, and the channel-level proposal and review tools), the script that splits and joins large files, and unit tests. The repository also contains the \texttt{ffprobe}-based video inventory, the FFmpeg scene-change screen (\texttt{tools/detect\_cuts.py}) and the web tool used for the manual review of candidate cuts (\texttt{tools/review\_cuts.py}) described in the Technical validation section, and the script that applies the review (\texttt{tools/apply\_cuts.py}): splitting segments and their detection files at confirmed cuts, and the scripts that remove short segments, merge duplicate locality records, and find and remove detection files that no longer match a segment. 

\bibliographystyle{unsrtnat} 
\bibliography{references}

\section*{Author Contributions}
Conceptualization: Md Shadab Alam, Pavlo Bazilinskyy;
Methodology: Md Shadab Alam, Pavlo Bazilinskyy;
Software: Md Shadab Alam, Pavlo Bazilinskyy;
Data curation: Md Shadab Alam, Olena Bazilinska, Pavlo Bazilinskyy;
Investigation: Md Shadab Alam, Pavlo Bazilinskyy;
Formal analysis: Md Shadab Alam;
Validation: Pavlo Bazilinskyy;
Visualization: Md Shadab Alam, Pavlo Bazilinskyy;
Writing-original draft: Md Shadab Alam;
Writing-review \& editing: Olena Bazilinska, Pavlo Bazilinskyy;
Supervision: Pavlo Bazilinskyy;
Project administration: Pavlo Bazilinskyy;
Funding acquisition: Md Shadab Alam, Pavlo Bazilinskyy.\\
All authors have read and agreed to the published version of the manuscript.

\section*{Competing Interests}
The authors declare that they have no known competing financial interests or personal relationships that could have appeared to influence the work reported in this paper.

\section*{Acknowledgements}
The authors thank Ying Fang, Margarida Fortes Ferreira and Aloysia Prakoso for their valuable assistance in sourcing and preparing the video material used in this work. The authors also gratefully acknowledge Linghan Zhang, the Digital Twin Lab (DT Lab) at Eindhoven University of Technology (\url{https://www.tue.nl/en/research/institutes/eindhoven-artificial-intelligence-systems-institute/digital-twin-lab}), and the TU/e Supercomputing Center (HPC) (\url{https://supercomputing.tue.nl/}) for providing access to their systems and computational resources used in this study.

\section*{Funding}
This work used the Dutch national e-infrastructure with the support of the SURF Cooperative using grant no. EINF-1603.

\end{document}